\documentclass[lettersize,journal]{IEEEtran}
\usepackage{xcolor,soul,framed} %,caption

\colorlet{shadecolor}{yellow}
\usepackage{color,soul}
\usepackage[pdftex]{graphicx}
\graphicspath{{../pdf/}{../jpeg/}}
\DeclareGraphicsExtensions{.pdf,.jpeg,.png}

\usepackage{array}
\usepackage{mdwtab}
\usepackage{eqparbox}
\usepackage[hyphens]{url}

\usepackage{algorithmic}
\usepackage{balance}
\usepackage{euler}
\usepackage{verbatim}
\usepackage{algorithm}
\usepackage{resizegather}
\usepackage{newpxmath}
\usepackage{subfigure}
\usepackage{graphicx}
\usepackage{caption}
\usepackage{adjustbox}
\usepackage{multirow}
\usepackage{tikz}
\usepackage{tikz-qtree}
\usepackage{hyperref}
\usepackage[numbers]{natbib}
\usepackage{booktabs}
\makeatletter
\newcommand{\Rmnum}[1]{\expandafter\@slowromancap\romannumeral #1@}
\makeatother
\usepackage{siunitx}

\begin{document}

% \title{Goal-guided Transformer-enabled Reinforcement Learning for Efficient Autonomous Navigation with Interpretable Scene Representation}

\title{Goal-guided Transformer-enabled Reinforcement Learning for Efficient Autonomous Navigation}

\author{
Wenhui~Huang,~\IEEEmembership{Student Member,~IEEE,}
Yanxin~Zhou,
Xiangkun~He,~\IEEEmembership{Member,~IEEE,}
and Chen~Lv,~\IEEEmembership{Senior Member,~IEEE}% <-this % stops a space
\thanks{W. Huang, Y. Zhou, X. He, and C. Lv are with the School of Mechanical and Aerospace Engineering, Nanyang Technological University, Singapore, 639798. (E-mail: wenhui001@e.ntu.edu.sg, yanxin001@e.ntu.edu.sg, xiangkun.he@ntu.edu.sg, lyuchen@ntu.edu.sg)}

\thanks{Corresponding author: Chen Lv. (E-mail: lyuchen@ntu.edu.sg)}
}

        % <-this % stops a space



\maketitle

\begin{abstract}
Despite some successful applications of goal-driven navigation, existing deep reinforcement learning (DRL)-based approaches notoriously suffers from poor data efficiency issue. One of the reasons is that the goal information is decoupled from the perception module and directly introduced as a condition of decision-making, resulting in the goal-irrelevant features of the scene representation playing an adversary role during the learning process. In light of this, we present a novel Goal-guided Transformer-enabled reinforcement learning (GTRL) approach by considering the physical goal states as an input of the scene encoder for guiding the scene representation to couple with the goal information and realizing efficient autonomous navigation. More specifically, we propose a novel variant of the Vision Transformer as the backbone of the perception system, namely Goal-guided Transformer (GoT), and pre-train it with expert priors to boost the data efficiency. Subsequently, a reinforcement learning algorithm is instantiated for the decision-making system, taking the goal-oriented scene representation from the GoT as the input and generating decision commands. As a result, our approach motivates the scene representation to concentrate mainly on goal-relevant features, which substantially enhances the data efficiency of the DRL learning process, leading to superior navigation performance. Both simulation and real-world experimental results manifest the superiority of our approach in terms of data efficiency, performance, robustness, and sim-to-real generalization, compared with other state-of-the-art (SOTA) baselines. The demonstration video (\url{https://www.youtube.com/watch?v=aqJCHcsj4w0}) and the source code (\url{https://github.com/OscarHuangWind/DRL-Transformer-SimtoReal-Navigation}) are also provided.
\end{abstract}


\section{Introduction}
\IEEEPARstart{R}{einforcement} Learning (RL) algorithms have significantly contributed to a wide range of domains over the past years, including but not limited to autonomous driving \cite{he2022robust, wu2022safe, huang2019learning}, unmanned ground vehicle (UGV) navigation \cite{pfeiffer2018reinforced, kahn2021land}, and computer games \cite{wurman2022outracing}. With the representative capability of handling high-dimensional states, recent RL algorithms, e.g., deep Q-learning (DQN) \cite{sae1, huang2022sampling}, deep deterministic policy gradient (DDPG) \cite{lillicrap2015continuous, sae2}, and soft actor-critic (SAC) \cite{haarnoja2018soft, haarnoja2018soft1} are increasingly adopted by the robotics community to address decision-making problems, especially for autonomous navigation. 

Conventional autonomous navigation methods that rely on prior knowledge of maps have been well-studied thanks to the Simultaneous Localization and Mapping (SLAM) technique \cite{yousif2015overview, huang2022potential}. In reality, however, such an approach significantly depends on the map's precision and might even fail in an unknown environment. Therefore, developing a simple mapless navigation strategy directly utilizing sensor input, such as laser scan \cite{cimurs2021goal, zhu2022hierarchical} or visual images \cite{zhu2017target, wu2021learn} is an emerging field that is garnering significant attention in current UGV research. Especially having the advantage of visual fidelity, depth image-based autonomous navigation has been intensively studied by several works \cite{xie2017towards, xie2020learning}. Similarly, segmentation images \cite{liu2020wasserstein} are often employed in mapless end-to-end navigation as well due to their powerful representative capability \cite{mousavian2019visual, hawke2020urban}. In order to approach a target position, the approaches mentioned above train their models in a goal-conditional learning manner \cite{codevilla2018end}, directly concatenating the physical goal information (i.e., goal location in polar coordinate) with the latent states from the perception system (i.e., convolutional neural network) and feed into subsequent networks. Despite various degrees of success, these methods decouple the goal information from the scene representation, leading to poor data efficiency. For instance, latent states from the goal information-less scene encoder may include certain mismatched features that are unnecessary for reaching a goal position and thus play an adverse role during the RL training process.

Self-attention-based approaches, especially Transformers \cite{vaswani2017attention}, have become the dominant model of choice in the natural language processing field. Motivated by adapting a standard Transformer architecture to images with the fewest modifications, a variant named Vision Transformer (ViT) \cite{dosovitskiy2020image} that can deal with image input is proposed in the computer vision community and has been applied to various domains, such as robotic manipulation \cite{hansen2021stabilizing} and autonomous driving \cite{kargar2022vision}. However, there has yet to be an existing work that develops ViT-enabled DRL algorithms to UGV for realizing mapless autonomous navigation, especially for goal-driven tasks. 

In light of this, we present a novel Goal-guided Transformer-enabled reinforcement learning (GTRL) approach by considering the physical goal states as an input of the scene encoder for guiding the scene representation to couple with the goal information and achieving efficient autonomous navigation. To realize a ViT architecture that treats both physical and visual states as the input, we propose a novel ViT variant with minimal modifications, which we call Goal-guided Transformer (GoT) for the rest of the paper, as the backbone of our perception system. Then, we instantiate a GoT-enabled actor-critic algorithm, namely GoT-SAC, for the decision-making system, receiving the goal-oriented scene representation from the perception system and generating decision commands for the UGV. To boost the data efficiency, we pre-train the GoT with expert priors and then learn the decision-making with the subsequent RL process. As a result, our method makes the scene representation more interpretable in terms of reaching the goal information, which is confirmed through qualitative and quantitative evaluations. Most importantly, such an approach motivates the scene representation to concentrate mainly on goal-relevant features, which substantially enhances the data efficiency of the DRL learning process, leading to superior navigation performance. Therefore, the proposed approach is an efficient DRL-based autonomous navigation method for UGV from the goal-driven task perspective. We summarize the main contributions of this paper as follows:

\begin{enumerate}
     \item A novel and Transformer architecture-based DRL approach, Goal-guided Transformer-enabled reinforcement learning (GTRL), is realized to achieve an efficient goal-driven autonomous navigation for the UGV.
    \item A novel Transformer architecture is proposed, which we call Goal-guided Transformer (GoT) in this paper, through minimal modifications of ViT to handle the multimodal input: physical goal states and visual states. Most importantly, the GoT enables the scene representation to concentrate mainly on the goal-relevant features, significantly enhancing the data efficiency of the DRL learning process from the goal-driven task perspective.  
    \item As for the practical contribution, we instantiated a concrete GoT-enabled DRL algorithm for the proposed method and validated it both in simulation and the physical world. The experimental results demonstrate the clear superiority of the proposed approach in data efficiency and performance compared with other SOTA baselines. Moreover, the investigation of goal-driven navigation in the unknown environment confirms our approach's robustness and sim-to-real transferability.
\end{enumerate}

\section{Related Works}

\noindent Due to the powerful representative capability to high dimensional states and superior data efficiency, DRL algorithms are gaining increasing attention among the robot community \cite{zhu2022survey}, especially for autonomous navigation of the UGV. Several works that infer the decision and control commands from laser scans have been proposed thanks to their robust transfer performance from the simulation to the real world. For instance, \cite{zhelo2018curiosity} trains the Asynchronous Advantage Actor-Critic (A3C) algorithm with intrinsic reward signals measured by curiosity to achieve mapless navigation. In \cite{dobrevski2018map}, the steering angle is discretized into seven actions and trained together with the forward commands by the Advantage Actor-Critic (A2C) algorithm, and the trained model is applied to real-world obstacle avoidance. Similarly, \cite{tai2017virtual} presents goal-driven mapless navigation based on an asynchronous Deep Deterministic Policy Gradient (DDPG) algorithm and successfully generalizes the learned model to the physical environment. In order to reduce the training time, \cite{pfeiffer2018reinforced} proposes to pre-train the Constrained Policy Optimization (CPO) algorithm with imitation learning (IL) and then continuously train it in the RL manner.

Nevertheless, laser scans cannot provide sufficient information to describe the environment in some cases \cite{wu2021learn}, and thus scholars turn their attention to visual sensors-based approaches. In \cite{xie2020learning}, a DDPG algorithm that considers depth images as input is employed to train the control policy by switching the different controllers. Similarly, work from \cite{tai2018socially} utilizes the same depth-based information but combines Behavior Cloning (BC) and Generative Adversarial Imitation Learning (GAIL) to demonstrate the enhanced performance of the social force-driven path planning. \cite{mousavian2019visual} presents a deep learning model consisting of the Convolutional Networks (ConvNets) and Long Short-Term Memory (LSTM) network to make a decision among seven commands based on semantic segmentation images in real-time. However, existing works mainly learn goal-driven tasks in a goal-conditional learning manner which is mentioned in \cite{codevilla2018end}. Though this work focuses on conditional imitation learning (CIL), the logic behind utilizing goal information is similar to the DRL-based methods, treating the physical goal information as a condition of the decision-making and directly concatenating to the latent states provided by the perception system. On the contrary, we consider the physical goal information as an input of the scene encoder, rather than a condition, to extract matching features w.r.t. the goal-driven autonomous navigation and improve the DRL data efficiency. 

ViT-based architecture has been a dominant choice not only for computer vision (CV) tasks but also for robotic research to achieve better scene representation and analysis. In \cite{hansen2021stabilizing}, the ViT is utilized as one of the encoders to measure the stability of their manipulation approach. Similarly, \cite{godoy2022electromyography} proposes a temporal multi-channel ViT to classify the hand motions for achieving better control of the bionic hands. To learn a more effective global context of the scene, \cite{kargar2022vision} presents a perception utilizing ViT instead of ConvNets architecture, handling with the birds-eye-view (BEV) images. The simulation results indicate that such an encoder can identify the significant surrounding cars for the ego car to learn a safe and effective policy in complex environments. Another ViT-based work related to Vehicle-to-Everything (V2X) is presented in \cite{xu2022v2x}. They propose a robust cooperative perception framework by means of building a holistic attention model, effectively integrating information across road users. Despite various degrees of success in the above domains, to our best knowledge, no current work develops ViT-enabled DRL
algorithms for UGV's mapless autonomous navigation, especially for goal-driven tasks. Despite the superior representation capability, ViT is insufficient for achieving goal-driven autonomous navigation, as our task requires the scene encoder to handle the multi-modality of sensors. Several works related to multimodal ViT have recently been proposed to address computer vision tasks and achieve SOTA performance \cite{girdhar2022omnivore, bachmann2022multimae}. The primary motivation of these approaches for employing multimodal inputs is that supplying the RGB images with rich complementary information \cite{zhang2022cmx}. In contrast, our approach focuses on leveraging the input's multi-modalities to filter out goal-irrelevant information rather than enriching it. We accomplish this by fusing RGB images with physical goal states at the input level, extracting significant features oriented towards the goal.

\section{Preliminaries}
\subsection{Reinforcement Learning}
The objective of goal-driven autonomous navigation is that infers the linear and angular velocity of the UGV from the input states, including images and goal information. We consider such a task as a standard Markov decision process (MDP) formulated by a tuple $<\mathcal{S, A, P, R}>$, where $\mathcal{S}$ is a set of states denoting the possible condition of the agent and environment, $\mathcal{A}$ represents action space, $\mathcal{P}$ models the transition of the environment, and $\mathcal{R}$ is the reward function evaluating the future overall payoff. At each time step t, the RL agent percepts the state  $s_{t} \in \mathcal{S}$ and executes an action $a _{t} \in \mathcal{A}$, receiving an immediate reward $r_t = \mathcal{R}(s_{t},a_{t}):\mathcal{S\times A}\rightarrow \mathbb{R}$, as well as next state $s_{t+1} \in \mathcal{S}$ based on the transition probability $\mathcal{P}(s_{t+1}|s_{t},a_{t}):\mathcal{S\times A}\rightarrow [0, 1]$. Usually, the RL agent selects an action based on a policy $a_t \sim \pi(\cdot|s_{t}):\mathcal{S \rightarrow A}$, which represents a probability distribution denoting the belief that the agent holds about its decision at each time step. The target of the RL agent is to maximize the discounted total return along the future from an initial state $s$, i.e., $V^{\pi}(s)$, denoted as:
\begin{equation}
\small
\begin{aligned}
    V^{\pi}(s) = \underset{s_{t} \sim \mathcal{P}}{\mathbb{E}}[\sum_{t=0}^{T}\gamma^{t} \cdot r_{t}], 
\end{aligned}
\label{state}
\end{equation}

\noindent where $V^{\pi}$ is called value function and $\gamma$ is the discounting factor constrained by $0 < \gamma \leq 1$. Similarly, the state-value function $Q^{\pi}$ based on the state $s_{t}$ and the action $a_{t}$ at time step t is defined as:
\begin{equation}
\small
\begin{aligned}
    Q^{\pi}(s_{t},a_{t}) &= r_{t} + \gamma \cdot \underset{s_{t+1} \sim \mathcal{P}}{\mathbb{E}}[V^{\pi}(s_{t+1})] \\
    &= r_{t} + \gamma \cdot \underset{s_{t+1} \sim \mathcal{P}, \ a_{t+1} \sim \pi}{\mathbb{E}}[Q^{\pi}(s_{t+1}, a_{t+1})].
\end{aligned}
\label{action-state}
\end{equation}

In the actor-critic method, an optimal policy $\pi^{*}$ can be obtained by maximizing the overall future payoff for all states along one trajectory. Additionally, an entropy term can be augmented to the objective to prevent the policy from trapping in the local optima in the early stage \cite{haarnoja2017reinforcement}:
\begin{equation}
\small
\begin{aligned}
   \underset{\pi}{max} \quad \underset{s_{t} \sim \mathcal{P}, 
 a_{t} \sim \pi}{\mathbb{E}} [\sum_{t=0}^{T} \gamma^{t}(r_{t} + \alpha \mathcal{H}(\pi(\cdot|s_{t})))]
\end{aligned}
\label{objective}
\end{equation}

\noindent where $\alpha$ is the temperature parameter that balances between overall future payoff and Shannon entropy of the policy.

% \subsection{Behavior Cloning}
\subsection{Deep Imitation Learning}
As one technique of behavior cloning method in imitation learning (IL) field, deep imitation learning (DIL) aims at directly mimicking the decision policy given a set of image state-action pairs $\mathcal{D} = \{<s_{i}, a_{i}>\}_{i=1}^{N}$, where N represents the number of samples. Therefore, it is a supervised learning problem by minimizing the statistical distance between action $a_{i}$ and parameterized function approximator $\mathbf{F}$$(s_{i};\psi)$:
\begin{equation}
\small
\begin{aligned}
    \underset{\psi}{minimize} \ \sum\limits_{i=1}^{N}\mathcal{L}(\mathbf{F}(s_{i};\psi), \ a_{i})
\end{aligned}
\label{bc}
\end{equation}

\noindent where $\mathcal{L}$ indicates loss function. Usually, we assume the action $a^{\mathbf{E}}$ is directly from a human expert which means the Eq. \ref{bc} can be reformulated as:
\begin{equation}
\small
\begin{aligned}
    \underset{\psi}{minimize} \ \sum\limits_{i=1}^{N}\mathcal{L}(\mathbf{F}(s_{i};\psi), \ a_{i}^{\mathbf{E}})
\end{aligned}
\label{expert}
\end{equation}

\begin{figure}[t]
  \begin{center}
  \includegraphics[width=3.5in]{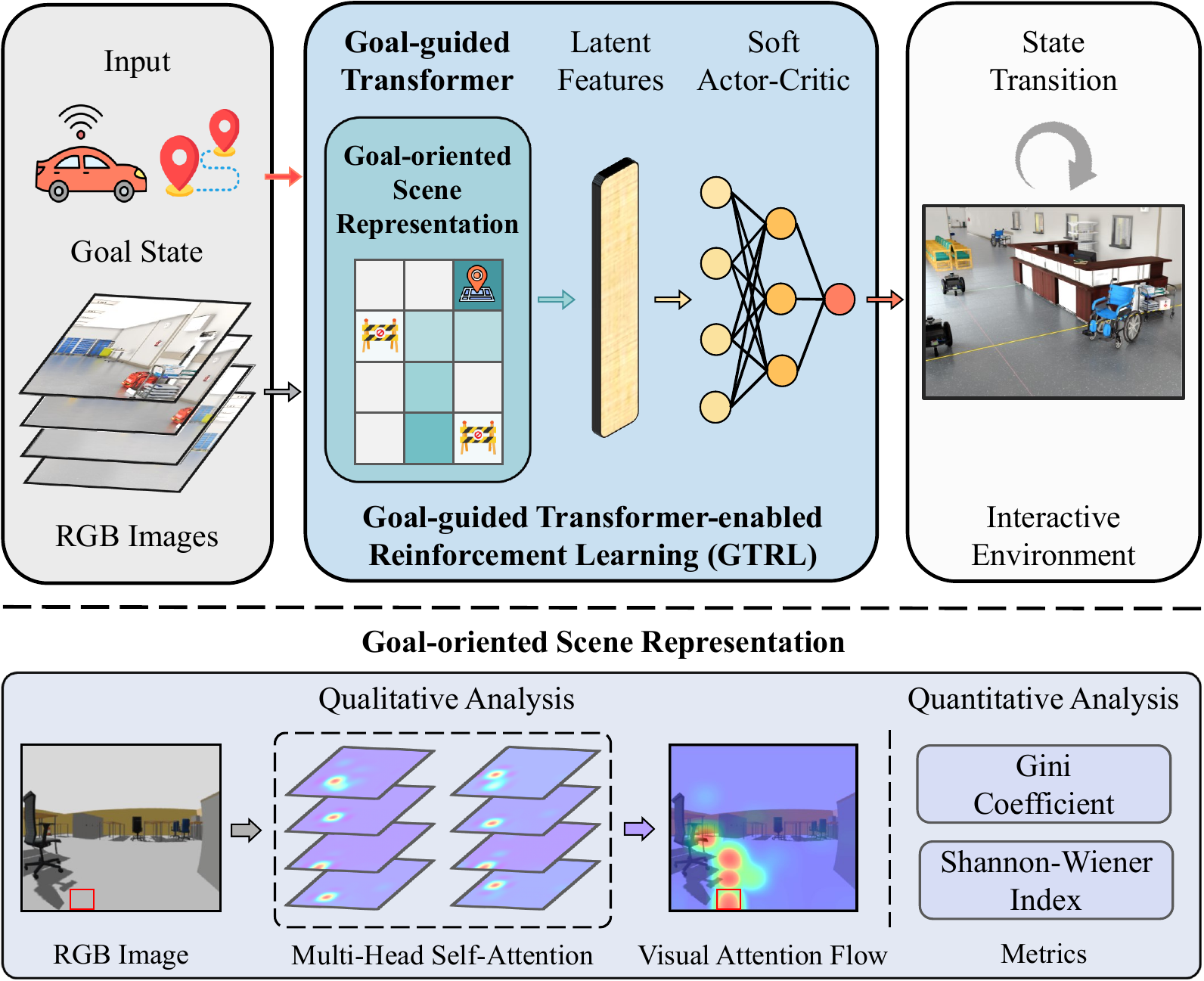}\\
  \caption{The overall framework of the proposed approach. The goal state in the polar coordination system is considered as input of the proposed approach through the entire learning process, guiding the scene representation to couple with the goal information and boosting the subsequent decision-making process. The goal-oriented scene representation is evaluated through both qualitative and quantitative analysis.}\label{framework}
  \end{center}
\end{figure}

\begin{figure*}[t]
  \begin{center}
  \includegraphics[width=6.7in]{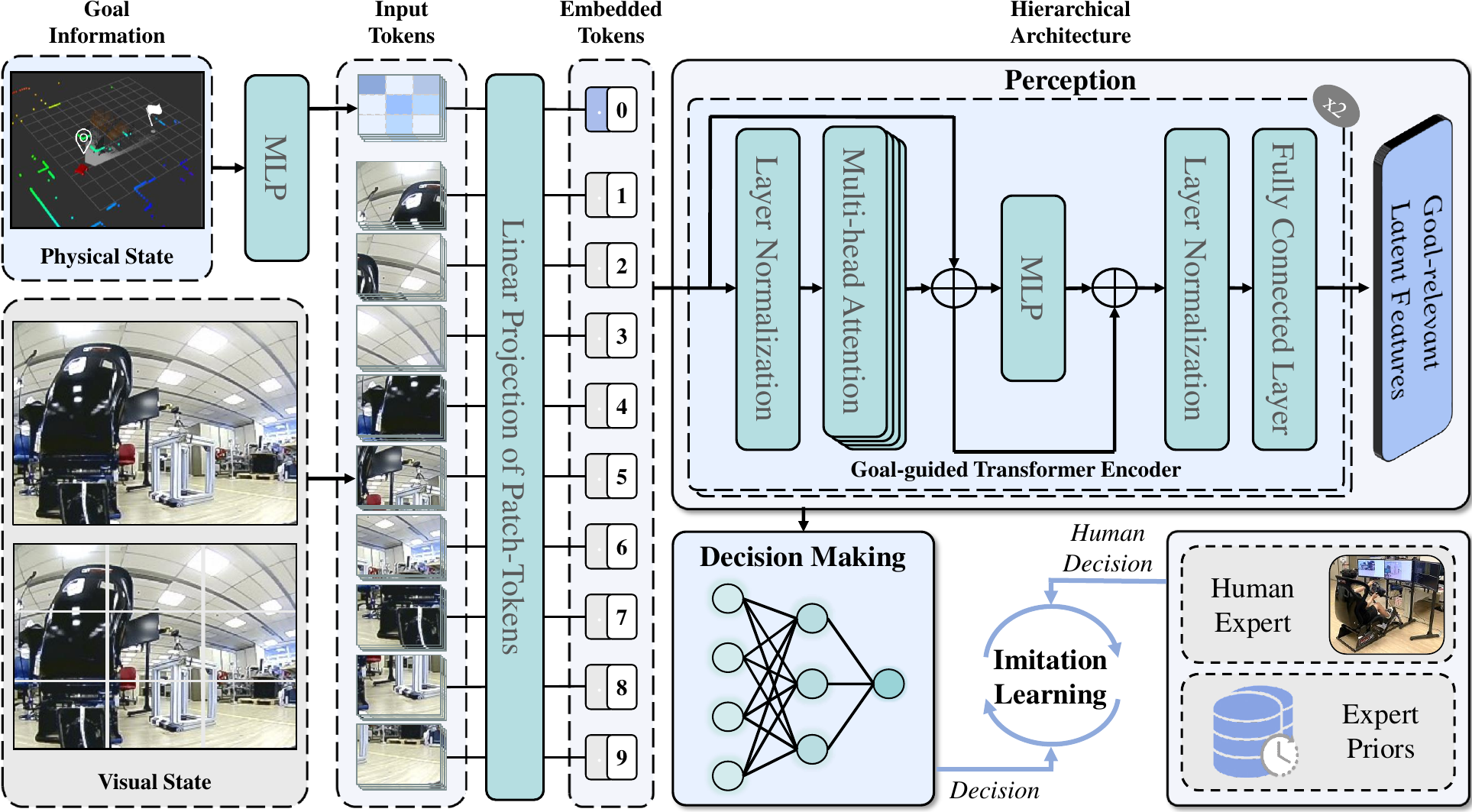}\\
  \caption{Goal-guided Transformer Architecture and Pre-train with Expert Priors. The physical goal state is encoded to the goal tokens; thus, the input tokens consist of both goal and visual information. Then the final inputs of the GoT are obtained by performing position embeddings on the input tokens and fed into GoT encoder. Next, the GoT extracts goal-relevant latent features by coupling the scene representation with the goal information and delivers them to the subsequent decision-making network. Finally, the GoT is pre-trained with expert priors through the DIL technique to boost data efficiency.}\label{architecture}
  \end{center}
\end{figure*}

\subsection{Vision Transformer}
The main idea behind ViT is splitting the images into patches and mapping them into linear embeddings in the same way the standard Transformer architecture treats tokens in natural language processing (NLP). Given an input image $x \in \mathbb{R}^{H \times W \times C}$, the ViT first reshapes it into a sequence of symbol representation ($x_{1}, x_{2}, ... x_{n}$), where ($H, W, C$) are the resolution and channel dimension of the input image $x$ and $x_{n} \in \mathbb{R}^{N \times (P^{2} \cdot C)}$ is a representation of flattened 2D patches with the resolution $P$. Therefore, the total number of 2D patches can be calculated as follows:
\begin{equation}
\small
\begin{aligned}
    N = \frac{H \cdot W}{P^{2}}
\end{aligned}
\label{number of patches}
\end{equation}

Then, the input of the ViT encoder can be obtained by augmenting the position embeddings $\mathbf{E}_{pos} \in \mathbb{R}^{(N+\mathit{1}) \times D}$ to D-dimensional flattened 2D patches:
\begin{equation}
\small
\begin{aligned}
    z_{0} = [x_{0};\ \mathbf{LP}(x_{1});\ \mathbf{LP}(x_{2}); \cdots ;\ \mathbf{LP}(x_{n})] + \mathbf{E}_{pos}
\end{aligned}
\label{input}
\end{equation}

\noindent where $\mathbf{LP}$ represents linear projection, and $x_{0} \in \mathbb{R}^{1 \times D}$ is an extra learnable embedding called class token. By feeding the embedded patches into the classic Transformer encoder, we can get multi-head self-attention (MSA) through the self-attention (SA) mechanism:
\begin{equation}
\small
\begin{aligned}
    MSA(Q,K,V) &= \mathbf{LP}([\mathbf{ATT}_{1}(Q,K,V);\ \mathbf{ATT}_{2}(Q,K,V);\cdots ;\ \mathbf{ATT}_{k}(Q,K,V)])
\end{aligned}
\label{MSA}
\end{equation}

\noindent where $k$ denotes k-th head and $\mathbf{ATT}$ indicates self-attention (SA) mechanism. As demonstrated in \cite{vaswani2017attention}, we compute SA through the query Q, keys K, and values V:
\begin{equation}
\small
\begin{aligned}
    \mathbf{ATT}(Q,K,V) &= softmax(\frac{QK^{T}}{\sqrt{d_{k}}})V \\
    [Q,K,V] &= \mathbf{LP}(z)
\end{aligned}
\label{ATT}
\end{equation}

\noindent where z represents a set of embedded patches and $d_{k}$ is a scaling factor.

\begin{figure*}[t]
  \begin{center}
  \includegraphics[width=5.5in]{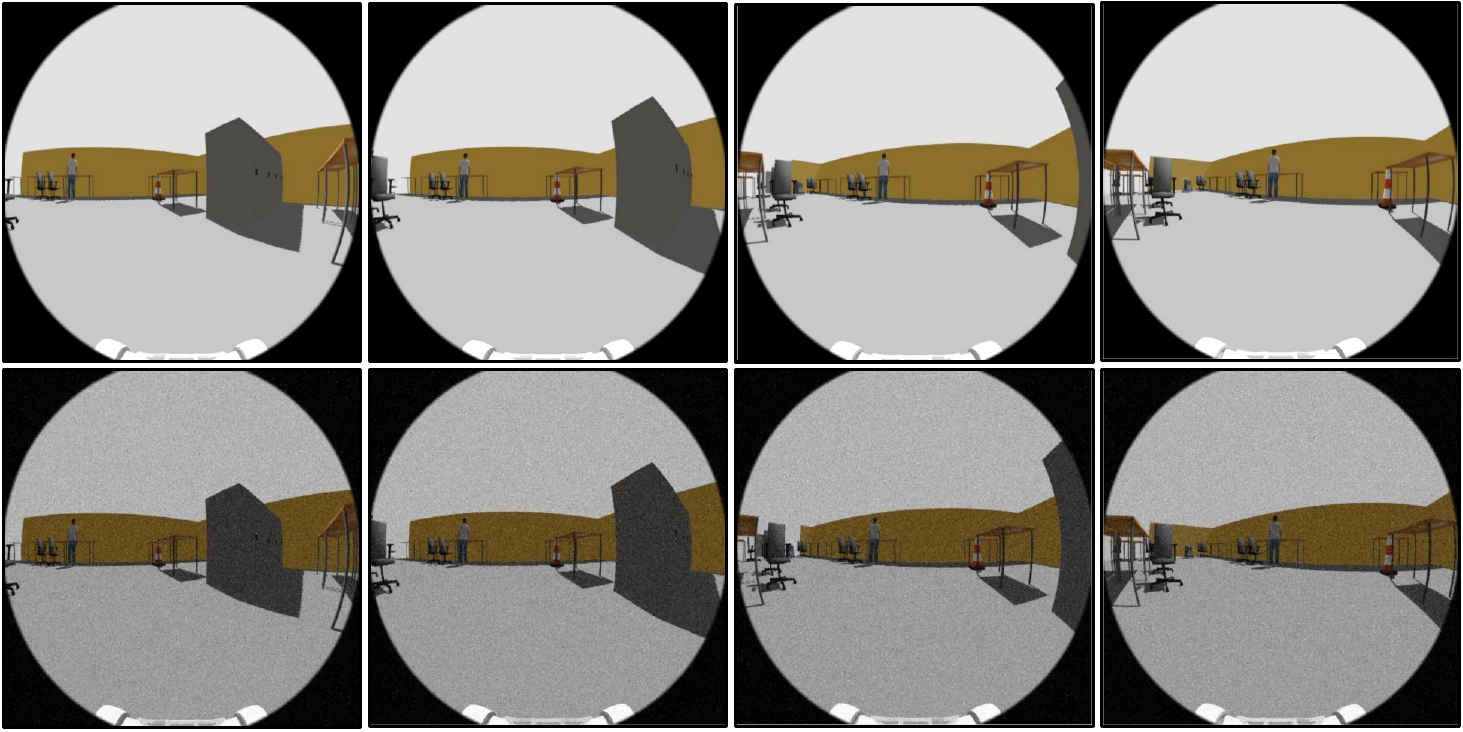}\\
  \caption{RGB images from the fisheye camera stacked for most recent four frames. The upside pair of figures show the raw RGB images, whereas those on the downside illustrate pixel-level Gaussian noise-augmented images after preprocessing.}\label{observation}
  \end{center}
\end{figure*}

\section{Methodology}
\subsection{Framework}\label{framework_approach}
The main aim of the goal-driven autonomous navigation task is to successfully navigate to a specific target instance within an unknown environment. Realizing such mapless navigation requires the DRL-based approach to understand and analyze the goal information. One possible solution is to treat the parameterized goal states as an input rather than a condition, feeding it together with the visual input, such as raw RGB images, to enhance the capability of the scene representation. Specifically, we learn the goal-oriented scene representation through a novel Transformer-based architecture that considers multimodal (i.e., physical goal states and visual states) input as a sequence of continuous representations. In light of this, we term the backbone of our perception system Goal-guided Transformer (GoT). Once the goal-oriented latent features are extracted, we motivate the SAC algorithm to learn the decision policy for approaching the goal position by interacting with the environment. Therefore, the two main ingredients, GoT and Transformer architecture-based SAC algorithm, complete our approach that we term Goal-guided Transformer-enabled reinforcement learning (GTRL).

The overall framework of our approach is depicted in Fig. \ref{framework}. In our case, the input consists of two parts, i.e., goal position in polar coordinates and raw fisheye RGB images stacked over four frames. In the first stage, they are flattened into the same dimension and fed into the GoT encoder. Then, these embedded patches are encoded to goal-relevant latent features through the MSA and provided to the subsequent decision-making system. Finally, the GoT-SAC algorithm makes a decision according to the goal-relevant latent features, and the UGV executes the decision command to trigger the state transition of the environment. After the algorithm converges, we qualitatively (visual attention flow maps) and quantitatively (Gini coefficient and Shannon-Wiener Index) evaluate the trained model in terms of the SA mechanism to analyze and interpret the significance of the goal-oriented scene representation (Section \ref{exp}).

\subsection{Goal-guided Transformer}
In order to deal with the multimodality of the input, i.e., the goal states and visual images, we propose a novel variant of the ViT that we term GoT in this paper. In model design, We construct the architecture of the GoT by the minimum modification of ViT for the purpose of a simple setup. Specifically, inspired by BERT \cite{kenton2019bert}, we define a special goal token $\mathcal{G} \in \mathbb{R}^{1 \times D}$ that is mapped from input goal states $s_{goal} \in \mathbb{R}^{1 \times 2}$ through a multilayer perception (MLP) network:
\begin{equation}
\small
\begin{aligned}
    \mathcal{G} = \mathbf{MLP}(s_{goal})
\end{aligned}
\label{goal_token}
\end{equation}

\noindent Therefore, the embeddings of GoT can be formulated as:
\begin{equation}
\small
\begin{aligned}
    z^{'}_{0} &= \mathbf{E}_{input}(s, \mathcal{G}) \\
    z_{0} &= z^{'}_{0} + \mathbf{E}_{pos}
\end{aligned}
\label{embeddings}
\end{equation}

\noindent where $\mathbf{E}_{input}$ and $\mathbf{E}
_{pos}$ represent input embeddings and position embeddings. By feeding the embeddings to the GoT encoder:
\begin{equation}
\small
\begin{aligned}
    z^{'}_{l} &= \mathbf{MSA}(\mathbf{LN}(z_{l-1})) + z_{l-1}\\
    z_{l} &= \mathbf{FC}(\mathbf{LN}(\mathbf{MLP}(z^{'}_{l}) + z^{'}_{l}))
\end{aligned}
\label{output}
\end{equation}

\noindent where l indicates the l-th block. We decide the depth of GoT as two blocks in this work, and hence, the latent features can be obtained from the output of the second block, denoted as:
\begin{equation}
\small
\begin{aligned}
    h = \boldsymbol{GoT}(s, \mathcal{G};\varphi)
\end{aligned}
\label{latent features}
\end{equation}

\noindent where $\varphi$ represents parameters of the GoT.

Figure \ref{architecture} illustrates an overview of the GoT architecture and pre-train process. As the figure shows, the input consists of two modalities: goal information as the physical state and raw RGB images as the visual state. The physical state is fed into $\mathbf{MLP}$ network and encoded as feature patches while the visual state is decomposed to eight by eight small image patches (we illustrate this process with three by three image patches in the figure due to limited space). Therefore, we can obtain complete input tokens by integrating both kinds of patches. Furthermore, we add position embeddings for each input token and fix the one encoded from goal information to the first position in particular. As for the GoT encoder, it consists of an MSA block, the $\mathbf{MLP}$, the fully connected layer ($\mathbf{FC}$), the layer normalization operation \cite{ba2016layer}, and the residual connections \cite{he2016deep}. Considering the limited computational power and lightweight design, we employ two blocks of the encoder with only four heads per block. Having the latent features from perception and the subsequent decision system, we are able to perform DIL through expert demonstration data to pre-train the GoT, enabling the hot-start initialization in the subsequential training process. In a standard DIL, in terms of the goal-driven end-to-end navigation problem, the function approximator depends on the environment state $s_{i}$ and goal state $s_{\{goal,\ i\}}$:
\begin{equation}
\small
\begin{aligned}
    \underset{\psi, \ \psi_{s}}{minimize} \ \sum\limits_{i=1}^{N}\mathcal{L}(\mathbf{F}(\mathbf{F}_{s}(s_{i};\psi_{s}),\ s_{\{goal,\ i\}};\psi), \ a_{i}^{\mathbf{E}})
\end{aligned}
\label{goal_condition}
\end{equation}

\noindent where $\psi$ and $N$ are the parameters of the function approximator and the number of samples. In our proposed approach, however, the goal state is no longer a condition but an input. Thus, the objective of goal-oriented imitation learning becomes:
\begin{equation}
\small
\begin{aligned}
    \underset{\psi}{minimize} \ \sum\limits_{i=1}^{N}\mathcal{L}(\mathbf{F}(\boldsymbol{GoT}(s_{i}, \mathcal{G}_{i});\psi), \ a_{i}^{\mathbf{E}})
\end{aligned}
\label{goal_orientation}
\end{equation}
%and provided to the decision system. 

In our case, such a design is essential since we aim to guide the scene representation to couple with the physical goal information so that the perception can extract goal-relevant and rational features to promote the data efficiency of the subsequent goal-driven decision process. To clearly demonstrate the point, we visualize goal-oriented scene representation through visual attention flow maps \cite{kim2017interpretable} and quantitatively evaluate the reliability of our approach in section \ref{exp}. Additionally, this design allows us to generalize the Transformer architecture to the multimodal input while keeping the original characteristics.

\begin{algorithm}[t]
\caption{Goal-guided Transformer-enabled Reinforcement Learning (GTRL)}
\begin{algorithmic}
\STATE {Initialize Goal-guided Transformer (GoT) network with pre-trained parameters: $\varphi^{*}$}.
\STATE {Initialize actor and critic network parameters: $\phi$, $\theta$}.
\STATE {Initialize entropy parameters: $\alpha$}.
\STATE{Initialize batch size N and replay buffer $\mathcal{D} \gets \varnothing$}.
\STATE {Assign target parameters: $\theta_{target} \gets \theta$}.
\STATE {$\mathbf{for}$ episode=1 to E do}
\STATE \hspace{0.2cm} Initialize the environment state: $s_{t} \sim Env$
\STATE \hspace{0.2cm} Initialize the goal state: $s_{\{goal,\ t\}} \sim Env$
\STATE \hspace{0.2cm} {$\mathbf{for}$ step=1 to S do}
\STATE \hspace{0.4cm} {Map goal token: $\mathcal{G}_{t} = \mathbf{MLP}(s_{\{goal, \ t\}})$}
\STATE \hspace{0.4cm} {Scene Representation:} 
\STATE \hspace{0.6cm} {$h_{t} \gets \boldsymbol{GoT}(s_{t}, \mathcal{G}_{t} \ ;\varphi^{*})$}
\STATE \hspace{0.4cm} {Sample an action: $a_{t} \gets \pi_{\phi}(a_{t}|h_{t})$}
\STATE \hspace{0.4cm} {Interact with the environment:} 
\STATE \hspace{0.6cm} {$r_{t}, s_{t+1}, s_{\{goal, \ t+1\}} \sim Env$}
\STATE \hspace{0.4cm} {Store the transition:} 
\STATE \hspace{0.6cm} {$\mathcal{D} \gets \mathcal{D} \cup (s_{t}, s_{\{goal, \ t\}}, a_{t}, r_{t}, s_{t+1}, s_{\{goal, \ t+1\}})$}
\STATE \hspace{0.4cm} {$\mathbf{If}$ time to update critic $\mathbf{then}$}
\STATE \hspace{0.6cm} {Sample a batch of the data:}
\STATE \hspace{0.8cm} {${(s_{t}^{i}, s_{\{goal, \ t\}}^{i}, a_{t}^{i}, r_{t}^{i}, s_{t+1}^{i}, s_{\{goal, \ t+1\}}^{i})}_{i=1}^{N} \sim \mathcal{D}$}
\STATE \hspace{0.6cm} {Compute critic (MBSE) loss: $\mathcal{L}(\theta)$ based on Eq. \ref{critic-loss}}.
\STATE \hspace{0.6cm} {Update parameters of critic network.}
\STATE \hspace{0.4cm} {$\mathbf{end}$ $\mathbf{if}$}
\STATE \hspace{0.4cm} {$\mathbf{If}$ time to update actor $\mathbf{then}$}
\STATE \hspace{0.6cm} {Sample a batch of the data:}
\STATE \hspace{0.8cm} {${(s_{t}^{i}, s_{\{goal, \ t\}}^{i}, a_{t}^{i}, r_{t}^{i}, s_{t+1}^{i}, s_{\{goal, \ t+1\}}^{i})}_{i=1}^{N} \sim \mathcal{D}$}
\STATE \hspace{0.6cm} {Compute actor loss: $\mathcal{L}(\phi)$ based on Eq. \ref{actor-loss}}.
\STATE \hspace{0.6cm} {Update parameters of actor network.}
\STATE \hspace{0.6cm} {$\mathbf{If}$ automatic tune is True $\mathbf{then}$}
\STATE \hspace{0.8cm} {Update temperature parameter $\alpha$.}
\STATE \hspace{0.6cm} {$\mathbf{end}$ $\mathbf{if}$}
\STATE \hspace{0.4cm} {$\mathbf{end}$ $\mathbf{if}$}
\STATE \hspace{0.4cm} {$\mathbf{If}$ time to update target network $\mathbf{then}$}
\STATE \hspace{0.6cm} {Update target network: $\theta_{target} \gets \theta.$}
\STATE \hspace{0.4cm} {$\mathbf{end}$ $\mathbf{if}$}
\STATE \hspace{0.2cm} {$\mathbf{end}$ $\mathbf{for}$}
\STATE {$\mathbf{end}$ $\mathbf{for}$}
\end{algorithmic}
\label{algo}
\end{algorithm}

\begin{figure}[t]
\centering
    \subfigure[Gazebo Environment.]{\includegraphics[width=.24\textwidth]{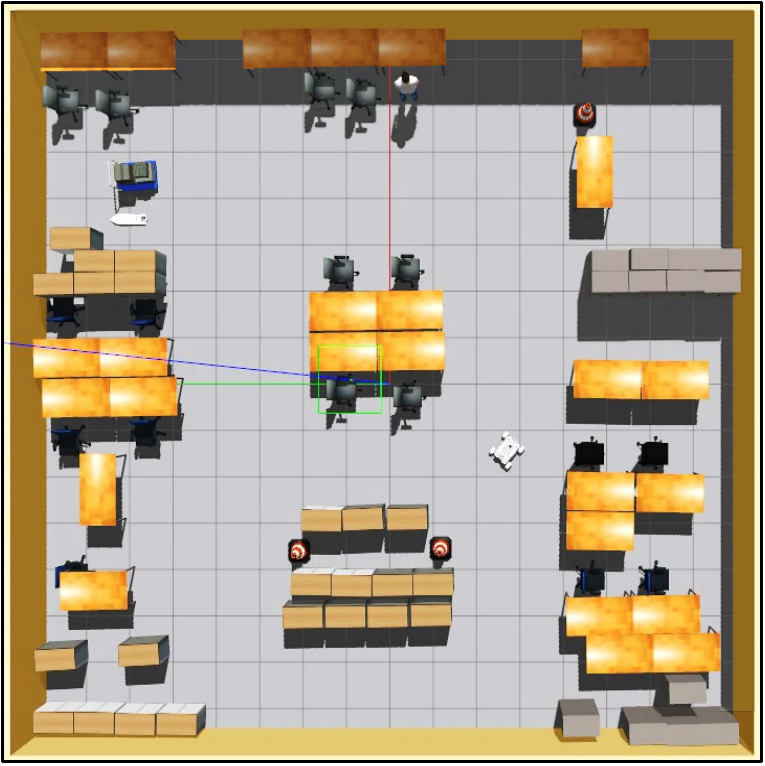}\label{simulation:a}}
    \subfigure[UGV.]{\includegraphics[width=.24\textwidth]{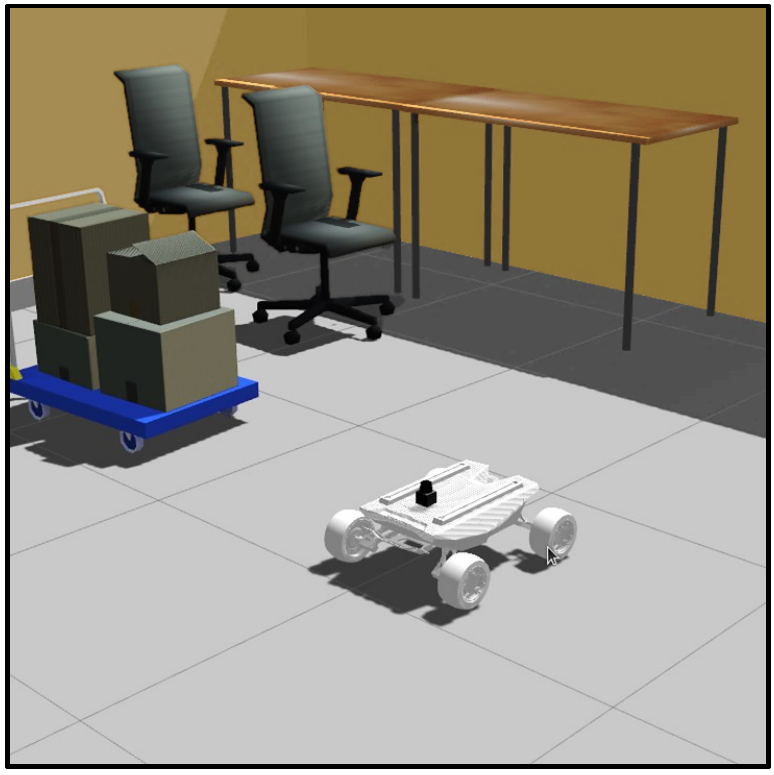}\label{simulation:b}}
\captionsetup{labelfont=bf}
\caption{Laboratory environment and UGV model.}
\label{simulation}
\end{figure}

\subsection{Goal-guided Transformer-enabled Reinforcement Learning}
As mentioned in section \ref{framework_approach}, the input of GTRL consists of two ingredients: visual states with raw RGB images and goal states in polar coordinates. In this work, we employ 160 $\times$ 120 raw RGB images from a fisheye camera with a FOV of 220 degrees and stack the four most recent frames. Additionally, we augment a pixel-level noise to the input images to learn a more robust and transferable decision policy for the sim-to-real experiments. Figure \ref{observation} demonstrates the difference between the original images and our input. The upside pair of four figures show the most recent four raw RGB images from the fisheye camera, whereas those on the downside illustrate Gaussian noise-augmented images that are utilized for training our algorithm. As for the goal state $s_{goal}$, we provide it in a 2-dimensional manner with the relative distance and heading error. Specifically, we define the first dimension of the goal state as the normalized relative distance and compute it as:
\begin{equation}
\small
\begin{aligned}
    d_{t} = \mathbf{min} (\frac{\Vert p^{<x,y>}_{t} - q^{<x,y>}\Vert_{2}}{\lambda},\ 1.0)
\end{aligned}
\label{goal_state_1}
\end{equation}

\noindent where $p^{<x,y>}_{t}$ denotes the real-time position of the UGV, $q^{<x,y>}$ indicates an arbitrary location of the goal point, $\Vert \cdot \Vert _{2}$ represents euclidean norm operation, and $\lambda$ is a constant normalizer that maps the relative distance in the range of [0, 1]. Correspondingly, we associate the second dimension of the goal state as the heading error between UGV's orientation and the directional vector points to the goal position:
\begin{equation}
\small
\begin{aligned}
  \Delta \varphi_{t} = \mathbf{atan}\big((q^{<y>} - p_{t}^{<y>}),\ (q^{<x>} - p_{t}^{<x>})\big) - \psi_{t}
\end{aligned}
\label{goal_state_2}
\end{equation}

\noindent where $\psi_{t}$ represents the heading angle of the UGV. 
Similar to the relative distance, we normalize the heading error as:
\begin{equation}
\small
\begin{aligned}
    \Delta \varphi_{t} =
    \begin{cases}
        \frac{\Delta \varphi_{t} - 2\pi}{\pi}, \quad if \ \Delta \varphi_{t} > \pi\\
        \frac{\Delta \varphi_{t} + 2\pi}{\pi}, \quad if \ \Delta \varphi_{t} < -\pi\\
        \frac{\Delta \varphi_{t}}{\pi}, \quad \quad \ otherwise
    \end{cases}
\end{aligned}
\label{goal_state_2_normalize}
\end{equation}

Receiving the above-mentioned input, the GTRL outputs decision commands $a_{t} = [v_{t}, \omega_{t}]$, i.e., linear velocity $v_{t} \in [0, 1]$ and angular velocity $\omega_{t} \in [-\frac{\pi}{2}, \frac{\pi}{2}]$, and delivers them to the UGV through the Robot Operating System (ROS). 

The target of autonomous navigation is to demonstrate a goal-driven decision and collision-free path planning for reaching the goal position. Therefore, we carefully design the reward function in combination with the continuous and sparse reward to boost the converge efficiency of the GTRL. More specifically, the overall payoff consists of four individual ingredients as follows:
\begin{equation}
\small
\begin{aligned}
    r_{t} = r_{h} + r_{a} + r_{g} + r_{c}
\end{aligned}
\label{reward}
\end{equation}

\noindent where $r_{h}$ denotes heuristic reward, $r_a$ represents action reward, $r_{g}$ indicates reward for arriving the goal position, and $r_c$ is the collision penalty. The heuristic reward is designed to motivate the UGV to move toward the goal position:
\begin{equation}
\small
\begin{aligned}
    r_{h} = \eta_{h} \times (\Vert p^{<x,y>}_{t-1} - q^{<x,y>}\Vert_{2} - \Vert p^{<x,y>}_{t} - q^{<x,y>}\Vert_{2})
\end{aligned}
\label{heuristic_reward}
\end{equation}

\noindent where $\eta_{h}$ is a constant weight. Similarly, we design the action reward to drive the UGV to approach the goal position as soon as possible but with the minimum number of steering operations:
\begin{equation}
\small
\begin{aligned}
    r_{a} = v_{t} - \eta_{a} \times \mathbf{abs}(\omega_{t})
\end{aligned}
\label{action_reward}
\end{equation}

\noindent where $\mathbf{abs}$ is an absolute value operation. Last but not least, two sparse rewards, i.e., the goal reach reward and the collision penalty, are designed as follows:
\begin{equation}
\small
\begin{aligned}
    r_{g} &= 
    \begin{cases}
        100, \quad \quad if \ d_{t} <= \xi \\
        0, \quad \quad \quad otherwise
    \end{cases} \\
    r_{c} &=
    \begin{cases}
        -100, \quad if \ collision\\
        0, \quad \quad \quad otherwise
    \end{cases}
\end{aligned}
\label{goal_reward}
\end{equation}

\noindent where $\xi$ represents a constant margin w.r.t. the goal position.

Subsequently, given the extracted latent features $h_{t}$ from GoT at a specific timestep t, the SAC algorithm learns the decision policy $\pi(a_{t}|h_{t})$ based on the reward function mentioned above. One common technique widely utilized in the SAC algorithm is double Q-networks to tackle the over-estimation issue. Hence, the parameters of the critic network of GoT-SAC are updated by minimizing the mean bellman-squared error (MBSE) loss function:
\begin{equation}
\small
\begin{aligned}
    \mathcal{L}(\theta_{i}) = \underset{h_{t} \sim \mathcal{P}, 
 a_{t} \sim \pi}{\mathbb{E}} \Vert Q_{\theta_{i}}^{\pi}(h_{t}, a_{t}) - (r_{t} + \gamma \cdot \hat{Q}^{\pi}) \Vert_{2}
\end{aligned}
\label{critic-loss}
\end{equation}

\begin{figure}[t]
  \begin{center}
  \includegraphics[width=3.25in]{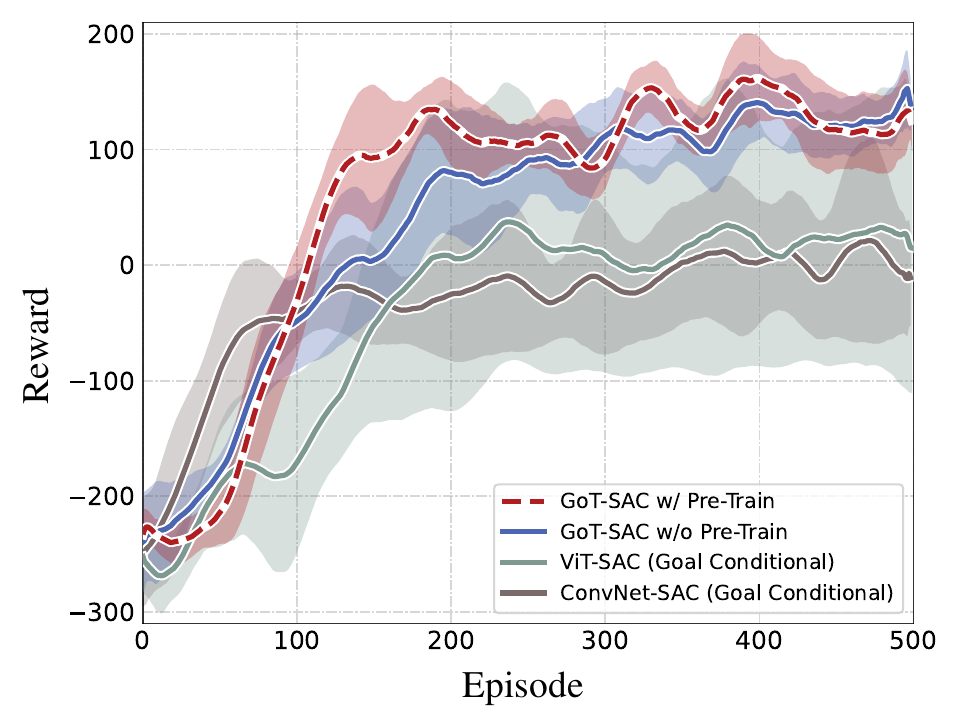}\\
  \caption{Convergence curve comparison. The red dotted line and solid lines represent the average rewards of our algorithms and baselines per episode, while the shaded areas depict the variances over five runs.}\label{reward_curve}
  \end{center}
\end{figure}

\noindent where $\hat Q^{\pi}$ is the state-action value of the next step from double target Q-networks and calculated by:
\begin{equation}
\small
\begin{aligned}
    \hat{Q}^{\pi} =\underset{h_{t+1} \sim \mathcal{P}, 
 a_{t+1} \sim \pi}{\mathbb{E}} \Big[\ \underset{i=1,2}{\mathbf{min}}Q_{\theta_{i}^{target}}^{\pi}(h_{t+1}, a_{t+1}) - \alpha \cdot log \pi(a_{t+1} | h_{t+1}) \Big]
\end{aligned}
\label{target-Q}
\end{equation}

\noindent where $\alpha$ is a temperature parameter that trades off between the stochasticity of the optimal policy and the state-action value. Accordingly, the actor network updates its parameters by maximizing the soft state-action function:
\begin{equation}
\small
\begin{aligned}
    \mathcal{L}(\phi) = \underset{h_{t} \sim \mathcal{P}, a_{t} \sim \pi}{\mathbf{E}} \Big[\underset{i=1,2}{\mathbf{min}}Q_{\theta_{i}}^{\pi}(h_{t},a_{t}) - \alpha \cdot log\pi_{\phi}(a_{t}|h_{t}) \Big]
\end{aligned}
\label{actor-loss}
\end{equation}

The detailed implementation of our approach is provided in Algorithm \ref{algo}.

\section{Experiments}\label{exp}

\subsection{Baseline Algorithms}
To benchmark the proposed GTRL method for trustworthy end-to-end autonomous navigation, we employ state-of-the-art RL and DIL algorithms as baselines to compare the qualitative and quantitative performance both in simulation and the real world.

\begin{figure}[t]
  % \vspace{0.1cm}
  \begin{center}
  \includegraphics[width=3.3in]{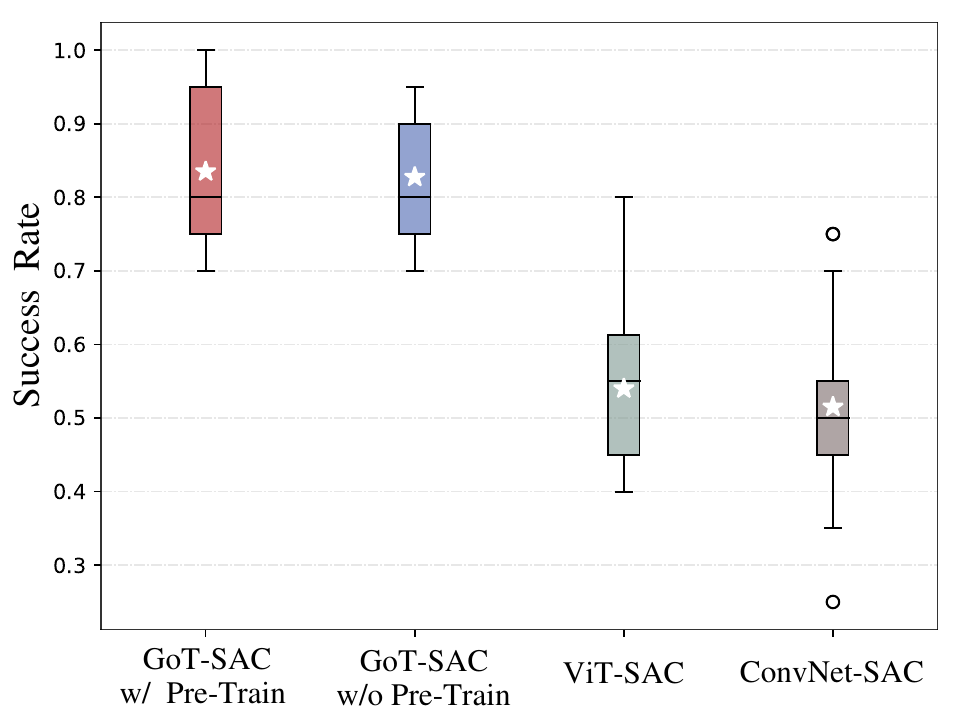}\\
  \caption{Success Rate Boxplot. The black-solid line and "star" located at the box body denote the median and average, while the hollow circles represent the outliers.}\label{boxplot}
  \end{center}
\end{figure}

\begin{figure*}[ht]
\centering
    \subfigure[Scenario \Rmnum{1}.]{\includegraphics[width=.328\textwidth, height=.22\textwidth]{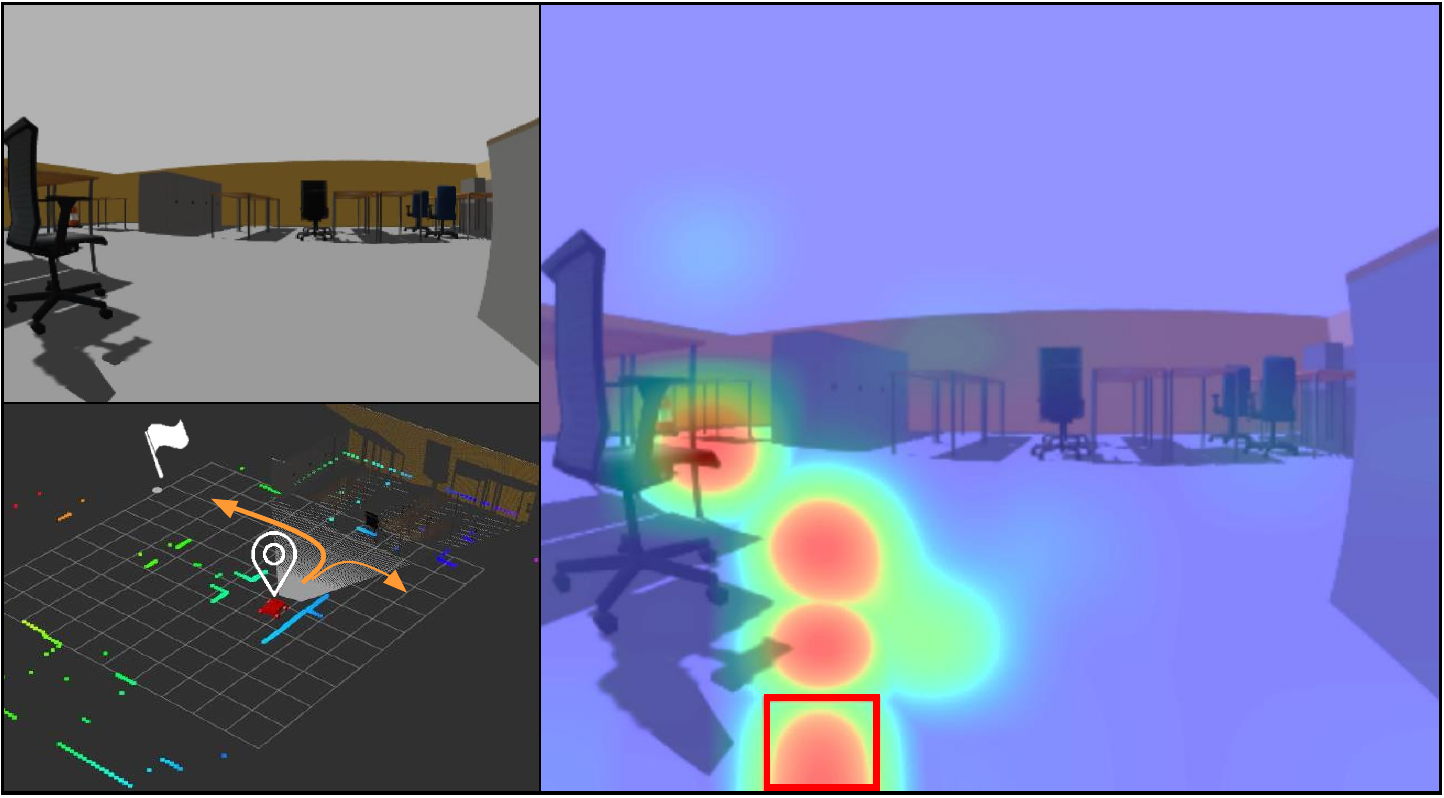}\label{attention:a}}
    \subfigure[Scenario \Rmnum{2}.]{\includegraphics[width=.328\textwidth, height=.22\textwidth]{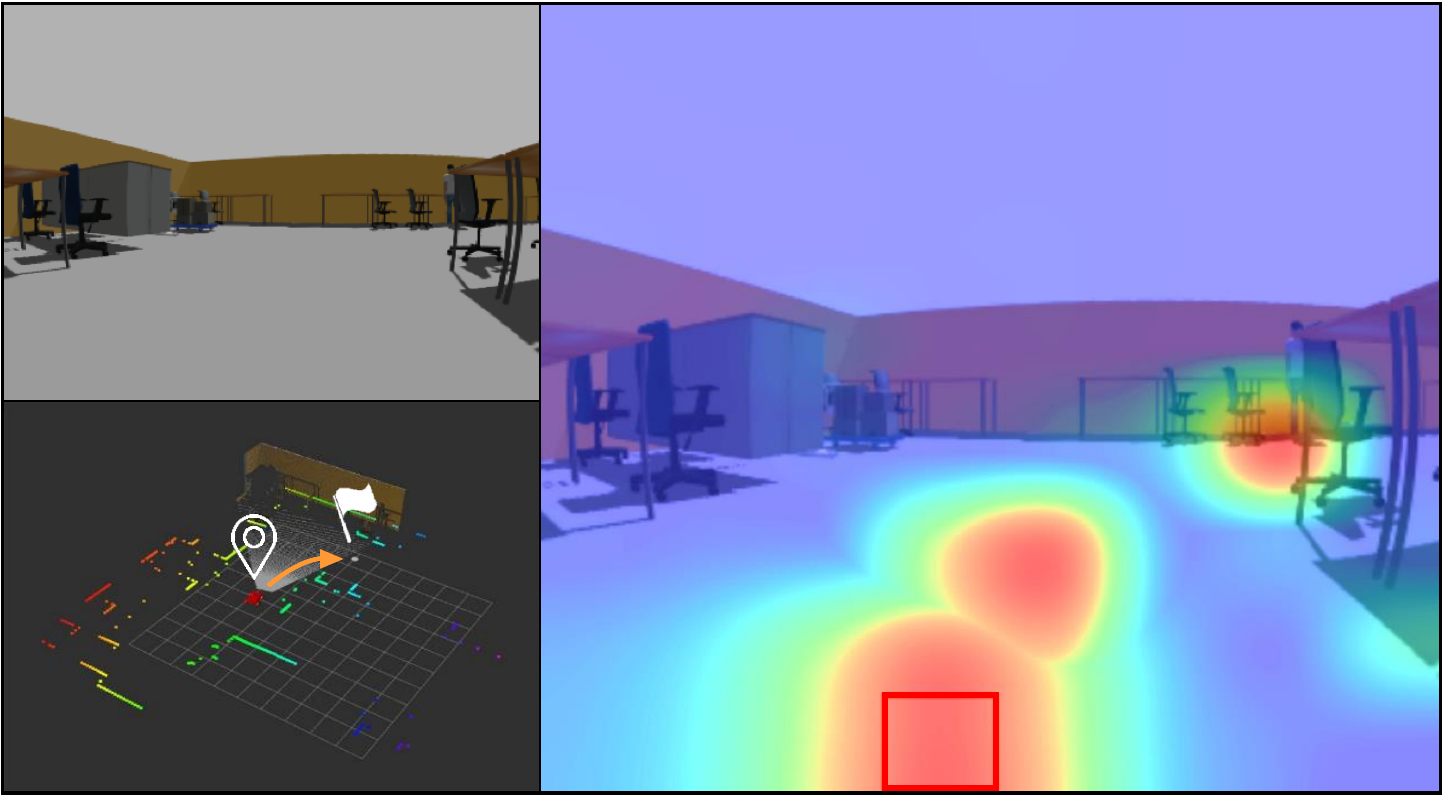}\label{attention:b}}
    \subfigure[Scenario \Rmnum{3}.]{\includegraphics[width=.328\textwidth, height=.22\textwidth]{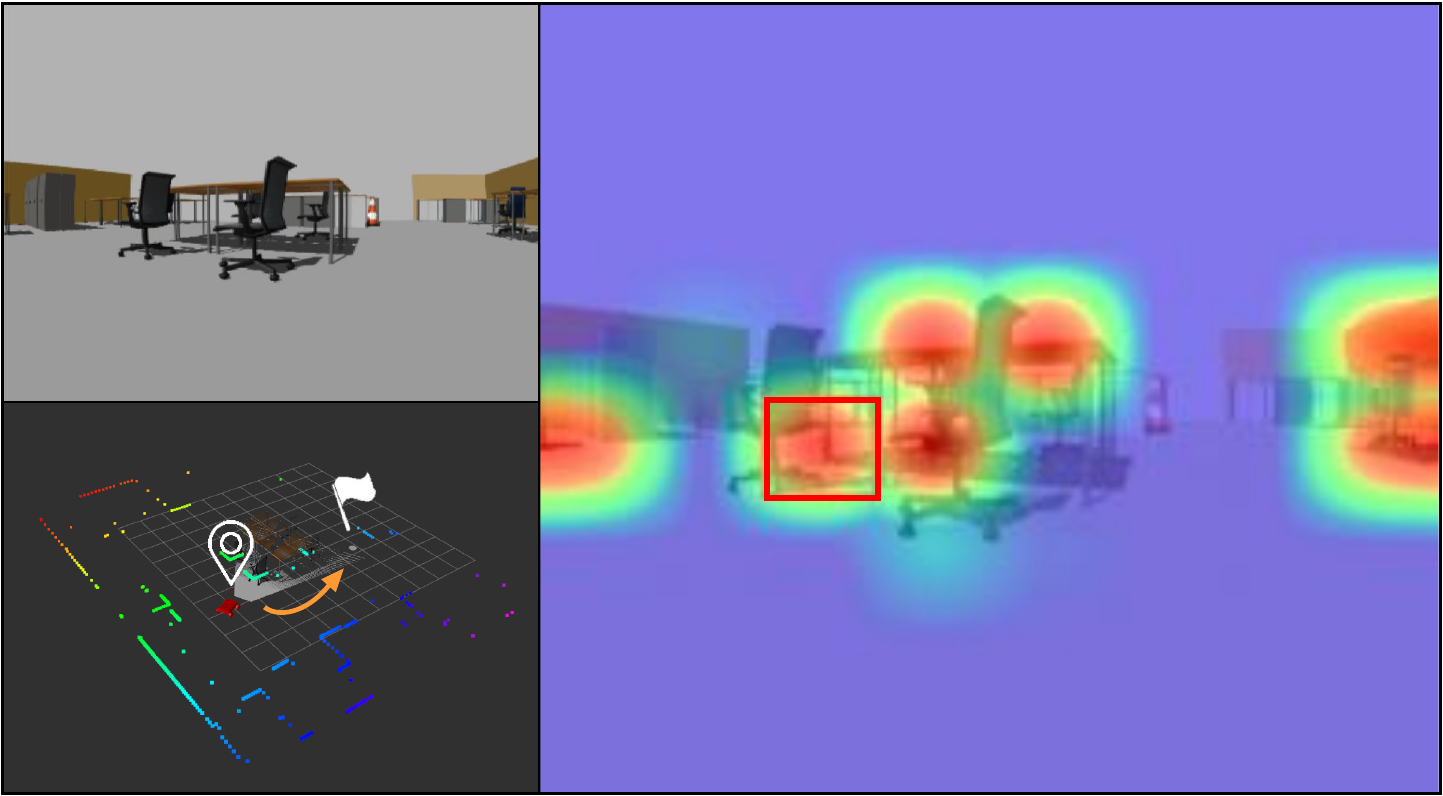}\label{attention:c}}
\captionsetup{labelfont=bf}
\caption{Attention Flow Visualization. The left pair of diagrams for each subfigure shows the original RGB image and goal information, while the right side diagram depicts the revised RGB image masked by the attention flow. A red square highlights the queried image patch, and the attention level is represented through a color transition from blue (low) to red (high). a) Scenario \Rmnum{1}: query for 59th image patch occupied with drivable space, b) Scenario \Rmnum{2}: query for 60th image patch occupied with drivable space, c) Scenario \Rmnum{3}:  query for 34th image patch occupied with obstacles.}
\label{attentionflow}
\end{figure*}

\begin{table*}[h]
\caption{Quantitative statistics of the self-attention mechanism behavior for the three goal-driven tasks. The data highlighted in bold denotes better results, i.e., higher Gini coefficient and lower Shannon-Wiener index, which depicts the attention are more concentrative on the significant image patches.}
\centering
\begin{adjustbox}{width=1.0\textwidth,center}
\begin{tabular}{c c c c c c c}
\toprule
\multirow{2}[3]{*}{Model} & \multicolumn{2}{c}{Episode I} & \multicolumn{2}{c}{Episode II} & \multicolumn{2}{c}{Episode III}\\
\cmidrule(lr){2-3} \cmidrule(lr){4-5} \cmidrule(lr){6-7}
 & Gini Coefficient & Shannon-Wiener Index & Gini Coefficient & Shannon-Wiener Index & Gini Coefficient & Shannon-Wiener Index\\
\midrule
\vspace{-0.2cm}
GoT-SAC & $\mathbf{0.927}$ & $\mathbf{0.896}$ & $\mathbf{0.848}$ & $\mathbf{1.133}$ & $\mathbf{0.901}$ & $\mathbf{0.984}$ \\
\\
ViT-SAC & 0.802 & 1.613 & 0.616 & 1.807 & 0.695 & 1.545\\
\bottomrule
\end{tabular}
\end{adjustbox}
\label{gini}
\end{table*}

\begin{enumerate}
    \item $\mathbf{ConvNet}$-$\mathbf{SAC}$ \cite{haarnoja2018soft}: A SOTA off-policy DRL algorithm that employs ConvNets as its scene representation encoder. We augment the physical goal state to the latent features encoded from ConvNets in a goal-conditional manner.
    \item $\mathbf{ViT}$-$\mathbf{SAC}$: This baseline is derived from a SOTA ViT-based DRL algorithm called ViT-DQN \cite{kargar2022vision}, which employs ViT-DINO as the backbone of the DQN encoder. Without losing the original vital characteristics, we replace the DQN with SAC to fit the end-to-end navigation demand and call it ViT-SAC in the rest of the paper. 
    \item $\mathbf{MultiModal}$ $\mathbf{CIL}$ \cite{xiao2020multimodal}: A SOTA conditional IL (CIL) algorithm that considers the human command or goal vector as a condition in the learning process. We select the command-input method among two architectures proposed in the original work to fit the goal-driven autonomous navigation task.
    \item $\mathbf{MoveBase \ Planner}$: A conventional planner widely utilized in UGV for goal-driven autonomous navigation. To be fair enough, we turn off the global map while keeping an eight-by-eight local map for real-time obstacle avoidance.
\end{enumerate}

In addition, we employ vanilla GoT-SAC to learn the policy from scratch without any expert priors during the reinforcement training process to validate our proposed algorithm's data efficiency thoroughly. 

\subsection{Expert Priors}

In order to pre-train the GoT, we asked the human participants to perform the demonstration in terms of the goal-driven navigation task and collected the data in image state-action pairs formation. During each episode, participants were given access to a randomly assigned goal position and were required to navigate toward it without collisions by continuously monitoring the fish-eye images displayed from a first-person perspective. More specifically, we provide the Logitech G29 driving set to the participants, allowing them to control the linear and angular velocity by manipulating the pedal/braking and steering wheel. To ensure the collection of reliable demonstrations, we selected two participants possessing valid driving licenses for the experiment. Additionally, a 10-minute training session was provided prior to the experiments to familiarize participants with the interaction devices and environments. Consequently, we obtained a total of 200 trajectories from human demonstrations, comprising 17,215 state-action pairs. These expert demonstrations were then divided into training and validation datasets, with a ratio of eight to two, resulting in 13,372 and 3,443 pairs, respectively, for use in the DIL process. The learning process terminates either when the maximum iteration limit is reached or when the validation loss turns to increase. The best model, determined by the lowest validation loss, is selected for the subsequent decision-making learning process.

\subsection{Simulation Assessment}
All algorithms are trained on a computer equipped with an Intel Core i7-10700 CPU, 64 GB of RAM, and an NVIDIA GTX 1660 SUPER graphics card. A high-fidelity autonomous navigation simulator, Gazebo, is employed to build the realistic laboratory environment and the UGV model for goal-driven mapless navigation, shown in Fig. \ref{simulation}. We train the instantiated algorithm for 500 episodes with a maximum of 200 steps for each. An episode ends when the goal position is reached, a collision occurs, or the UGV runs out of maximum step numbers. To well generalize the DRL-based policy and achieve better sim-to-real transferability, we not only augment a pixel-level Gaussian noise to the RGB image but also vary the initial location and goal position for each episode. Though the proposed algorithm only needs one fisheye camera for autonomous navigation, we also set a laser sensor in the simulation to detect the collision (\ref{simulation:b}). Furthermore, we employ the Robot Operating System (ROS) open-source platform to communicate with Gazebo and derive the goal information through subscribing to odometry messages.

Figure \ref{reward_curve} illustrates the learning curves of GoT-SAC and all the DRL-based baselines. We run each algorithm with five different random seeds to measure statistics and evaluate the robustness. Specifically, the red dotted line and solid lines represent the average rewards of our algorithms and baselines per episode, while the shaded areas depict the variances over five runs. As the figure shows, both versions of GoT-SAC achieve higher reward levels with relatively lower variances than those of goal-conditional DRL-based algorithms, which indicates the significance of the goal-oriented scene representation. Moreover, both GoT-SAC models exhibit a faster convergence and enhance the training efficiency by over 129\% and 86\% compared with the ViT-SAC model. It should be noticed that though the convergence pace of ConvNet-SAC at the early stage is slightly faster due to its relatively small number of parameters, the average episode return is much lower than our proposed algorithm. Overall, the convergence curve confirms the better data efficiency of the proposed approach, that is, achieving an enhanced performance (a higher reward level) by consuming less amount of the data (fewer training episodes) compared to other baselines.

To evaluate the performance, we validate all the trained policies with 20 random seeds and run for 50 episodes for each seed. The success rate, which is obtained as the number of goal-reached episodes divided by the total runs, is employed as the metric for the evaluation. From Fig. \ref{boxplot}, we can observe the dominant success rate and superior robustness of the proposed algorithm compared with other baselines regardless of varying a wide range of random seeds.

\subsection{Attention Visualization and Evaluation}
\begin{figure}[t]
  \vspace{0.1cm}
  \begin{center}
  \includegraphics[width=3.4in]{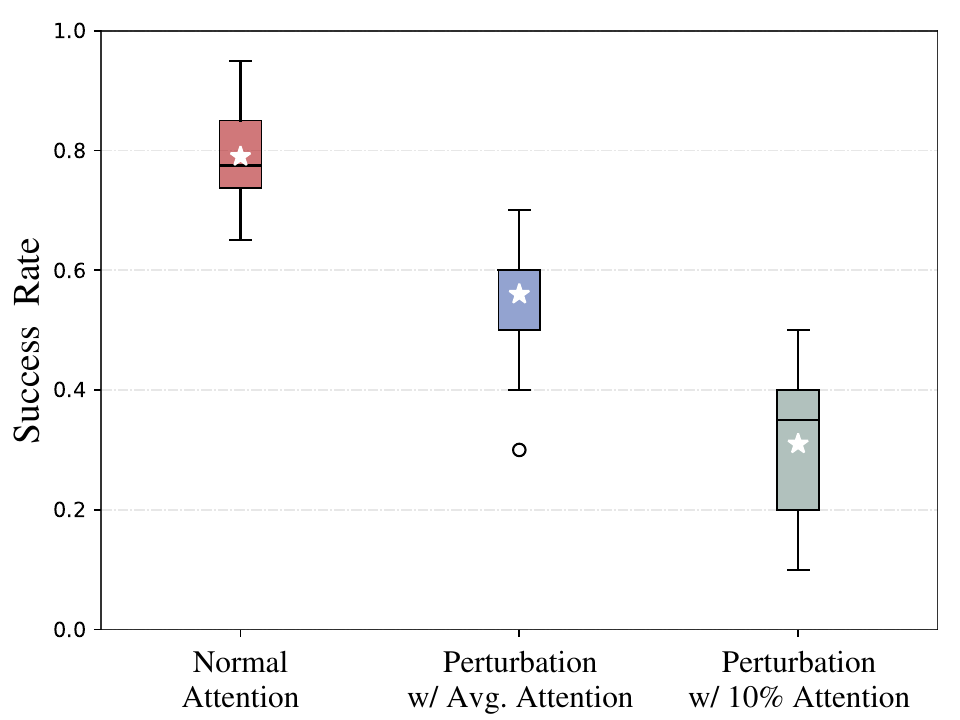}\\
  \caption{Success Rate Comparison in terms of the different attention levels. The black-solid line and "star" located at the box body denote the median and average, while the hollow circles represent the outliers.}\label{perturbation}
  \end{center}
\end{figure}

Besides the superior efficiency and performance, the GTRL approach also possesses a significant advantage in model interpretability thanks to the goal-oriented scene representation. To analyze the rationale behind fast convergence and excellent performance of our algorithm, we extract the attention from the GoT encoder w.r.t. randomly sampled RGB images and visualize it in Fig. \ref{attentionflow}. As the figure shows, the left pair of diagrams for each subfigure shows the original RGB image and goal information, while the right side diagram depicts the visual attention flow map \cite{kim2017interpretable} masked by the extracted attention. The queried image patch is highlighted by a red square, and the attention level is represented through a color transition from blue (low) to red (high). In Scenario \Rmnum{1} (Fig. \ref{attention:a}), the UGV is facing the oncoming T intersection, and the goal position locates on the left side behind the office chair and table. From the visual attention flow map, we can observe that the overall attention generates a visual path by mainly focusing on goal-oriented image patches. It should be noticed that the orientation of such a visual path obviously towards the goal position though the right turn is also feasible in this scenario, which proves that the scene representation successfully couples with the goal information. Similarly, a clear goal-driven visual path is shown in Scenario \Rmnum{2} (Fig. \ref{attention:b}). In Fig. \ref{attention:c}, different from the previous two scenarios, we query for the image patch that occupies an obstacle (office chair) instead of drivable space. We surprisingly find that the attention highlights most of the adjacent obstacles, evidently pointing out the undrivable regions. Therefore, we qualitatively verify that our approach can provide a clear explanation of how the UGV analyzes the scene and arrives at the destination with a collision-free path.

\begin{figure}[t]
  \vspace{0.1cm}
  \begin{center}
  \includegraphics[width=3.4in]{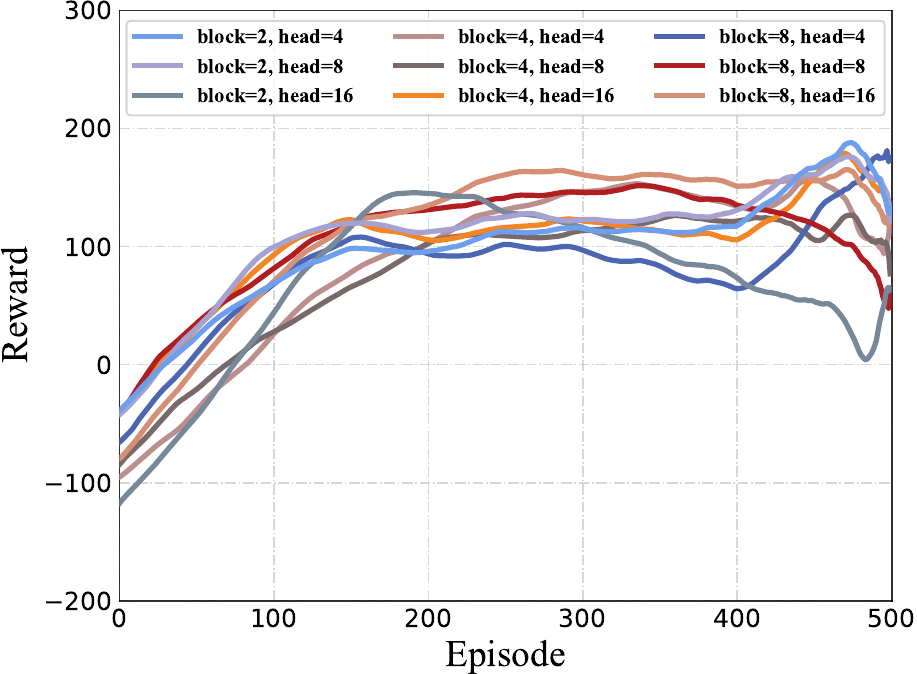}\\
  \caption{Convergence curve for different combinations of SA head and encoder block parameters.}\label{parameters}
  \end{center}
\end{figure}

\begin{table*}[t]
\caption{Computational efficiency of selected parameter settings.}
\centering
\begin{adjustbox}{width=1.0\textwidth,center}
\begin{tabular}{c c c c c c}
\toprule
\multicolumn{1}{c}{\multirow{2}[2]{*}{Settings}} & \multicolumn{1}{c}{\multirow{2}[2]{*}{Algorithm}} & \multicolumn{4}{c}{Complexity Metrics} \\
\cmidrule(lr){3-6}
& & FLOPs (M) & Params (M) & Inference (ms) & Train (hr) \\
\midrule
% \vspace{-0.3cm}
\multirow{2}[1]{*}{2 Blocks \& 4 Heads} & \multicolumn{1}{c}{GoT-SAC} & 
36.9 & 0.46 & 0.24 $\pm$ 0.02 & 3.16 $\pm$ 0.1\\
& \multicolumn{1}{c}{ViT-SAC} & 
36.2 & 0.45 & 0.23 $\pm$ 0.03 & 3.10 $\pm$ 0.15\\
\bottomrule
\end{tabular}
\label{complexity}
\end{adjustbox}
\end{table*}

Furthermore, we quantitatively evaluate and compare our approach with the ViT-SAC (goal-conditional) model to support the above conclusion. Specifically, we run each model with three random episodes and measure the statistical characteristics by averaging the whole frame. Realizing that this is an unsupervised task since there do not exist labels or ground truth for comparing, we employ two unsupervised metrics: Gini coefficient for measuring the evenness of the attention weights distribution \cite{letarte2018importance} and Shannon-Wiener index for evaluating the concentration of the attention \cite{kim2020attentional} in this experiment and the results are reported in Table \ref{gini}. The Gini coefficient and Shannon-Wiener index are widely utilized metrics for measuring statistical dispersion intended to represent the evenness and diversity of distribution. From the table, we can observe that both metrics clearly reveal that the attention of the GoT-SAC model is sparser and tends to be more concentrated on task-related image patches, proving that better interpretability is achieved through goal-oriented scene representation.

Last but not least, we also investigate the impact of the significant attention through perturbation-based method \cite{zablocki2022explainability} to observe how modifications of critical attention affect the navigation task performance. In light of this, we measure the success rate of the GoT-SAC model over another twenty random seeds with fifty episodes for each by dynamically replacing the essential attention (weights higher than 0.995) with a moving average and 10\% of the original value. The boxplot illustrated in Figure \ref{perturbation} shows the overall result. We can observe that the performance of the GoT-SAC model, the one that decreases the significant attention to 10\%, degrades catastrophically (62.5\%) in terms of the success rate. Though the success rate of the model employing the average perturbation method is slightly higher than the previous one, it is still clearly lower than the normal GoT-SAC model, indicating the significance of the attention learned by our approach.

\subsection{Ablation Study of Goal-guided Transformer Parameters}

\begin{figure}[t]
\centering
    \subfigure[SCOUTMINI.]{\includegraphics[width=.24\textwidth, height=.184\textwidth]{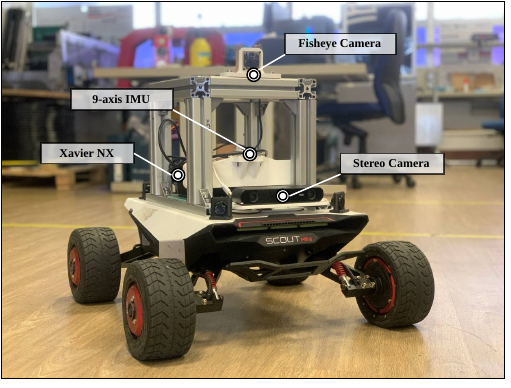}\label{scout}}
    \subfigure[Robotics Research Center.]{\includegraphics[width=.22\textwidth]{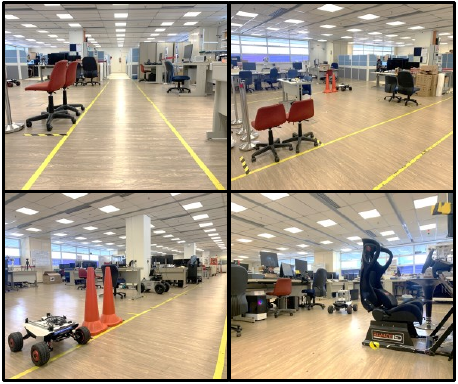}\label{rrc}}
\captionsetup{labelfont=bf}
\caption{The real UGV platform and sim-to-real experiment environment. a) SCOUTMINI: an omnidirectional steering mobile robot from Agilex. b) Robotics research center: an indoor laboratory space in Nanyang Technological University.}
\label{sim2real:environment}
\end{figure}

Acknowledging the significance of the attention mechanism discussed in the preceding section, it is of value to examine the influence of the architecture of the GoT, specifically the quantity of self-attention (SA) heads and encoder blocks. Thus, here we analyze the training performance of the GTRL in terms of the abovementioned two parameters. We fixed a random seed and tested the nine combinations of SA head and encoder block, shown in Fig. \ref{parameters}. From the figure, we can observe that all of the combinations successfully converge at a similar pace, though the overall performance slightly grows up as the number of SA heads and encoder blocks increases. More specifically, the group of eight blocks and sixteen heads demonstrates the fastest convergence speed and achieves the highest reward performance among nine combinations, while the worst one originated by the setting of two blocks and sixteen heads. Considering the principle of lightweight design and limited computation power of the hardware platform in the sim-to-real experiments, we employ two blocks of the encoder with four heads per block in this work. Table \ref{complexity} illustrates the quantitative comparison of computational efficiency between GoT-SAC and ViT-SAC algorithms with selected parameters.

\subsection{Sim-to-Real Assessment}
In addition to evaluating the feasibility and performance of the algorithm in a virtual simulation environment, we also expect to apply our approach in real-world navigation tasks. In terms of the UGV, we use the omnidirectional steering mobile platform from Agilex called SCOUTMINI. The SCOUTMINI equips with an edge computing platform NVIDIA Jetson Xavier, a ZED2i stereo camera, an inertial measurement unit (IMU), and a fisheye camera with an ultra-wide FOV of 220 degrees (Fig. \ref{scout}). Regarding the software, we deliver the goal information and raw fisheye RGB images to the UGV through ROS. Then, the GoT-SAC sends the real-time decision inference to the UGV chassis via controller area network (CAN) communication to realize motion control. In this real-world experiment, all the algorithms are applied at the Robotics Research Center at Nanyang Technological University to complete a loop navigation task, as shown in Fig. \ref{rrc}. More specifically, we design four destinations that motivate the UGV to reach one by one with a small break after each arrival and finally return to the vicinity of the starting point. This experiment aims to test the algorithm's ability to avoid static obstacles and quickly navigate to given goal positions.
\begin{figure*}[ht]
\centering
    \subfigure[GoT-SAC.]{\includegraphics[width=.18\textwidth, height=.21\textwidth]{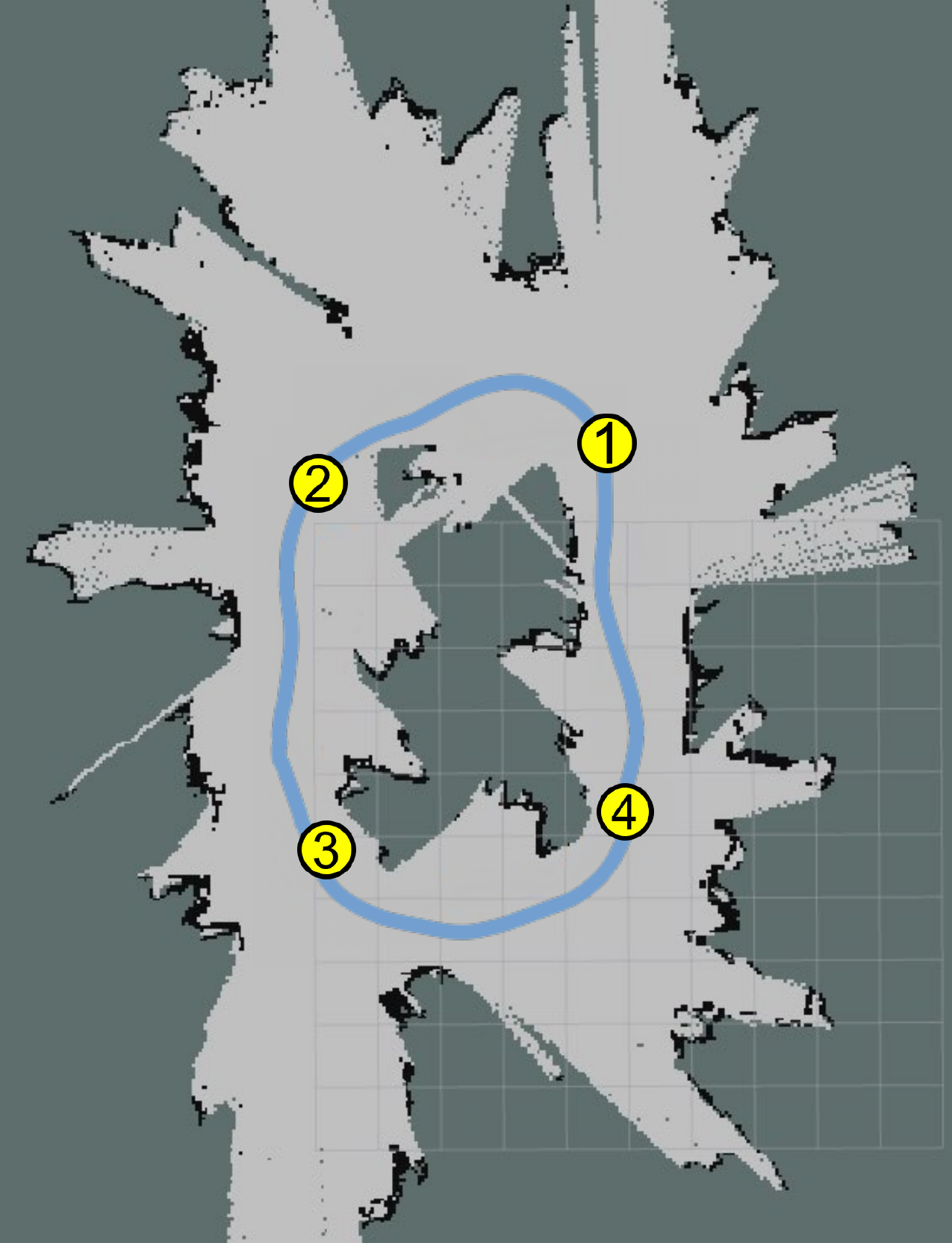}\label{scenario:a}}
    \subfigure[ViT-SAC.]{\includegraphics[width=.18\textwidth, height=.21\textwidth]{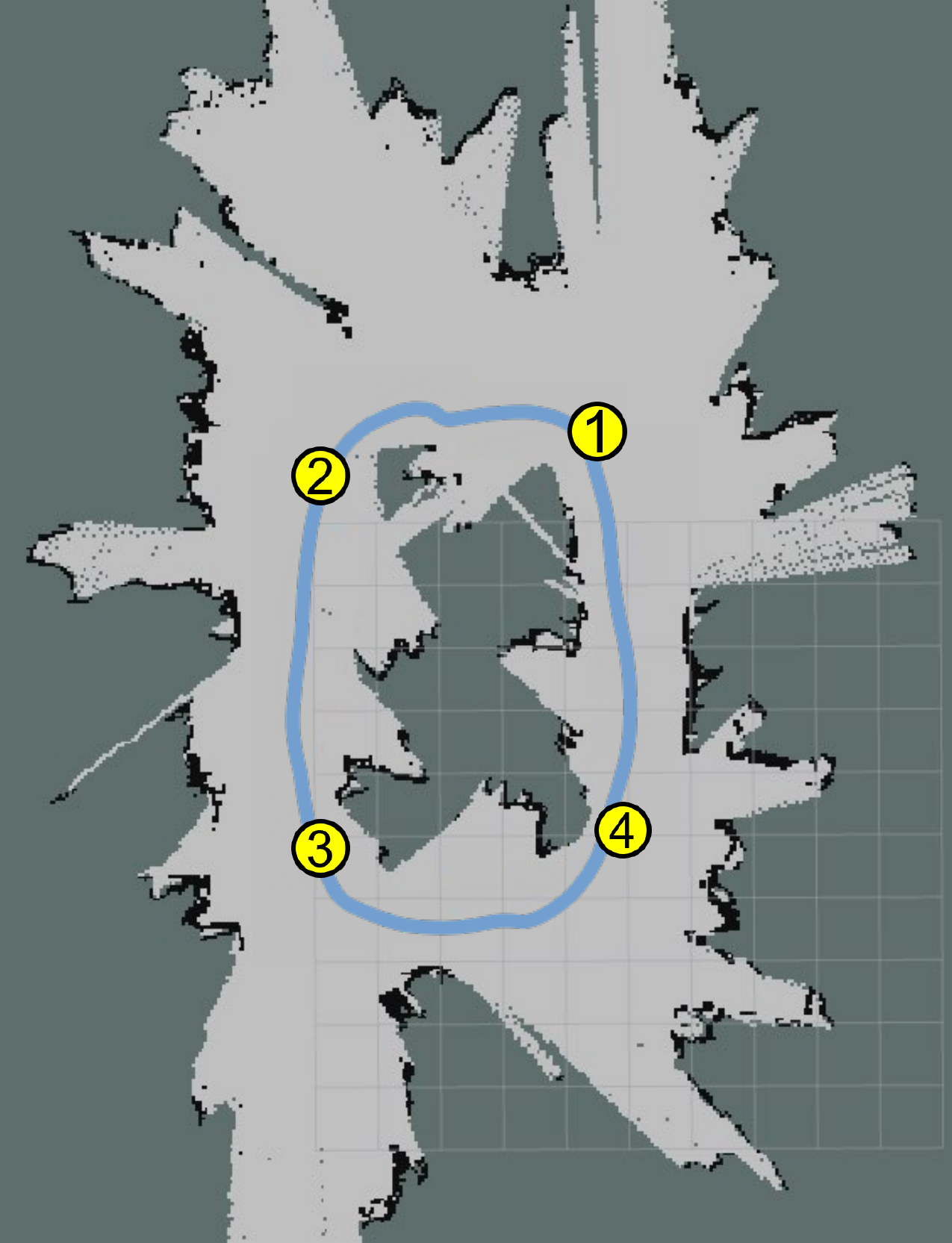}\label{scenario:b}}
    \subfigure[ConvNet-SAC.]{\includegraphics[width=.18\textwidth, height=.21\textwidth]{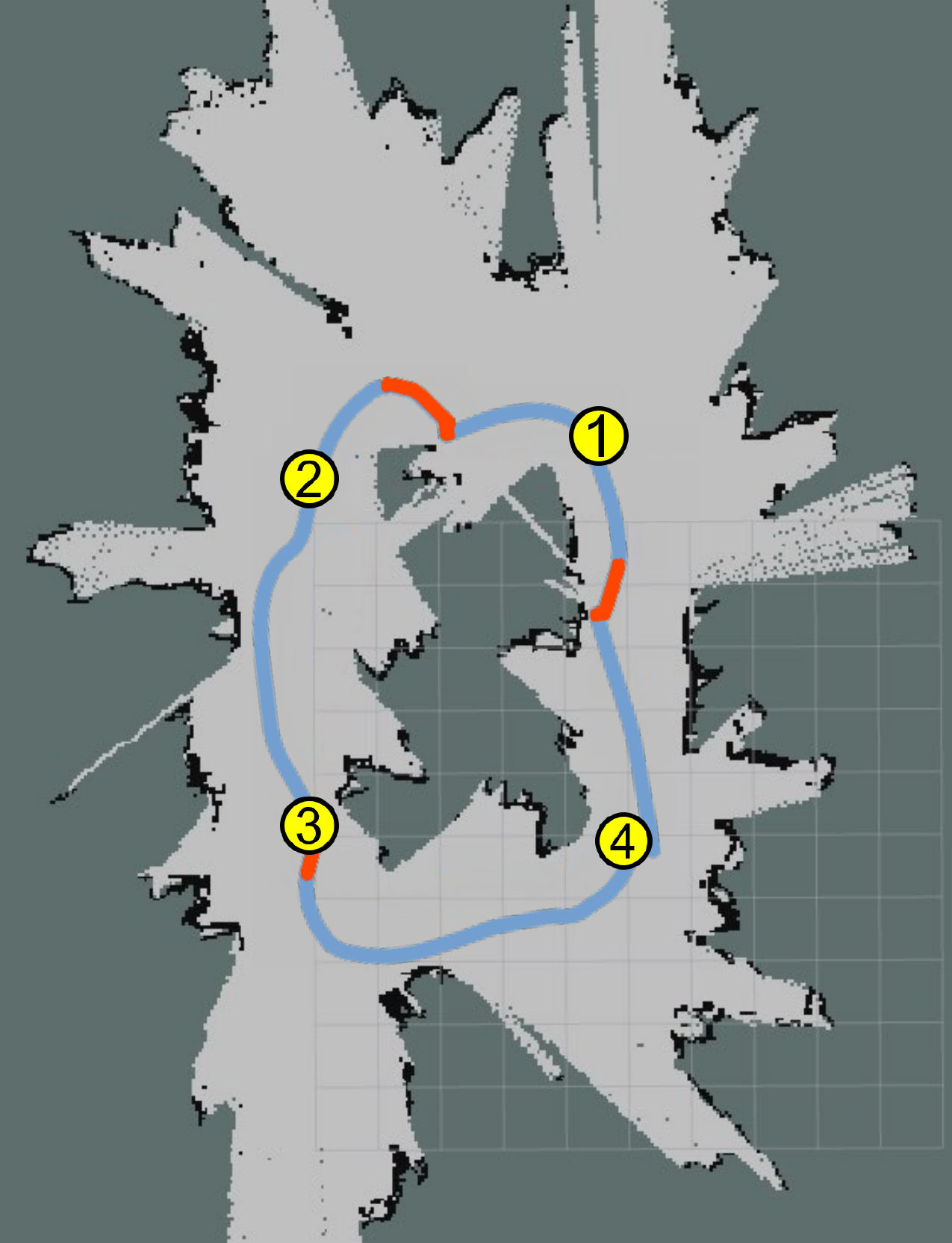}\label{scenario:c}}
    \subfigure[MultiModal CIL.]{\includegraphics[width=.18\textwidth, height=.21\textwidth]{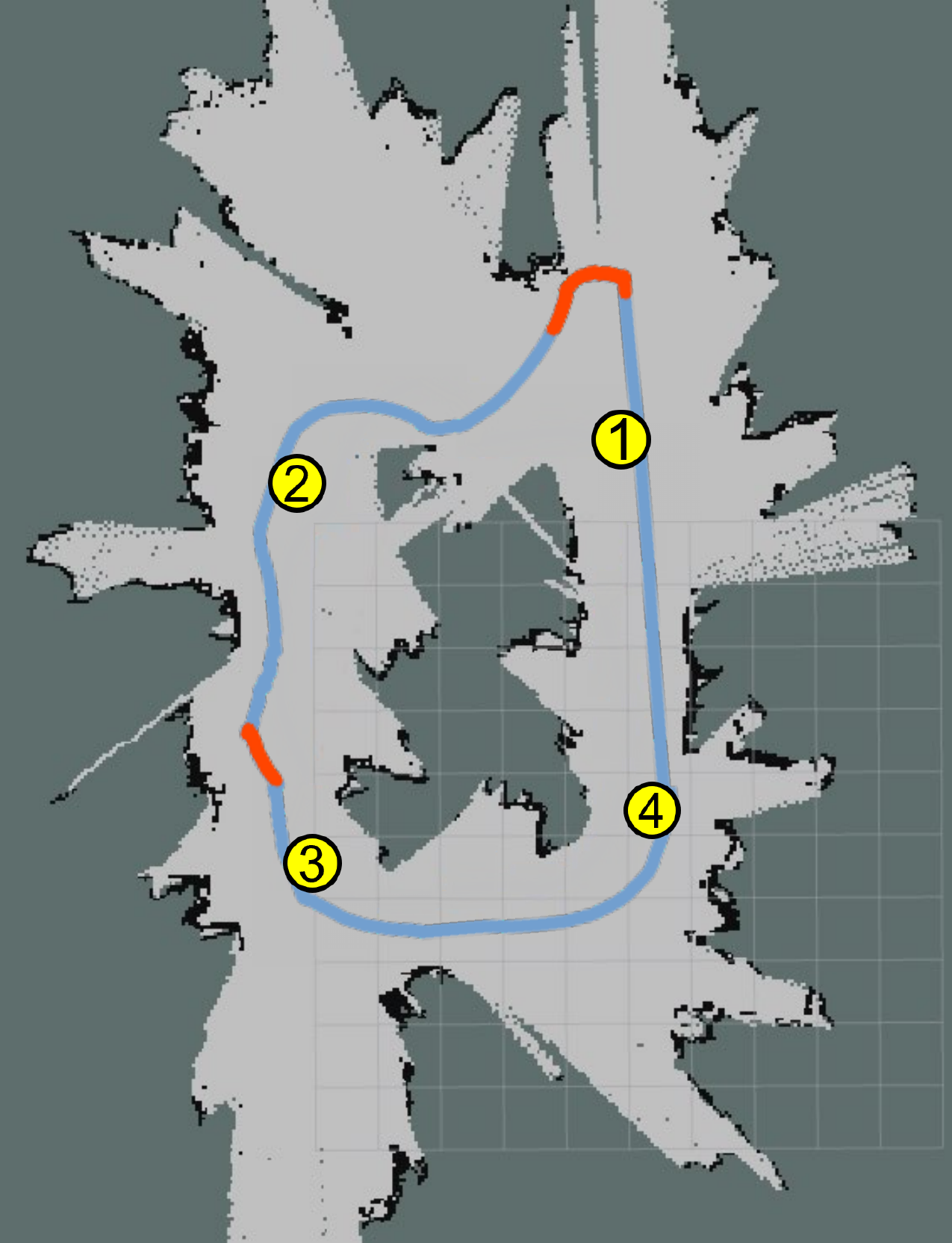}\label{scenario:d}}
    \subfigure[MoveBase.]{\includegraphics[width=.18\textwidth, height=.21\textwidth]{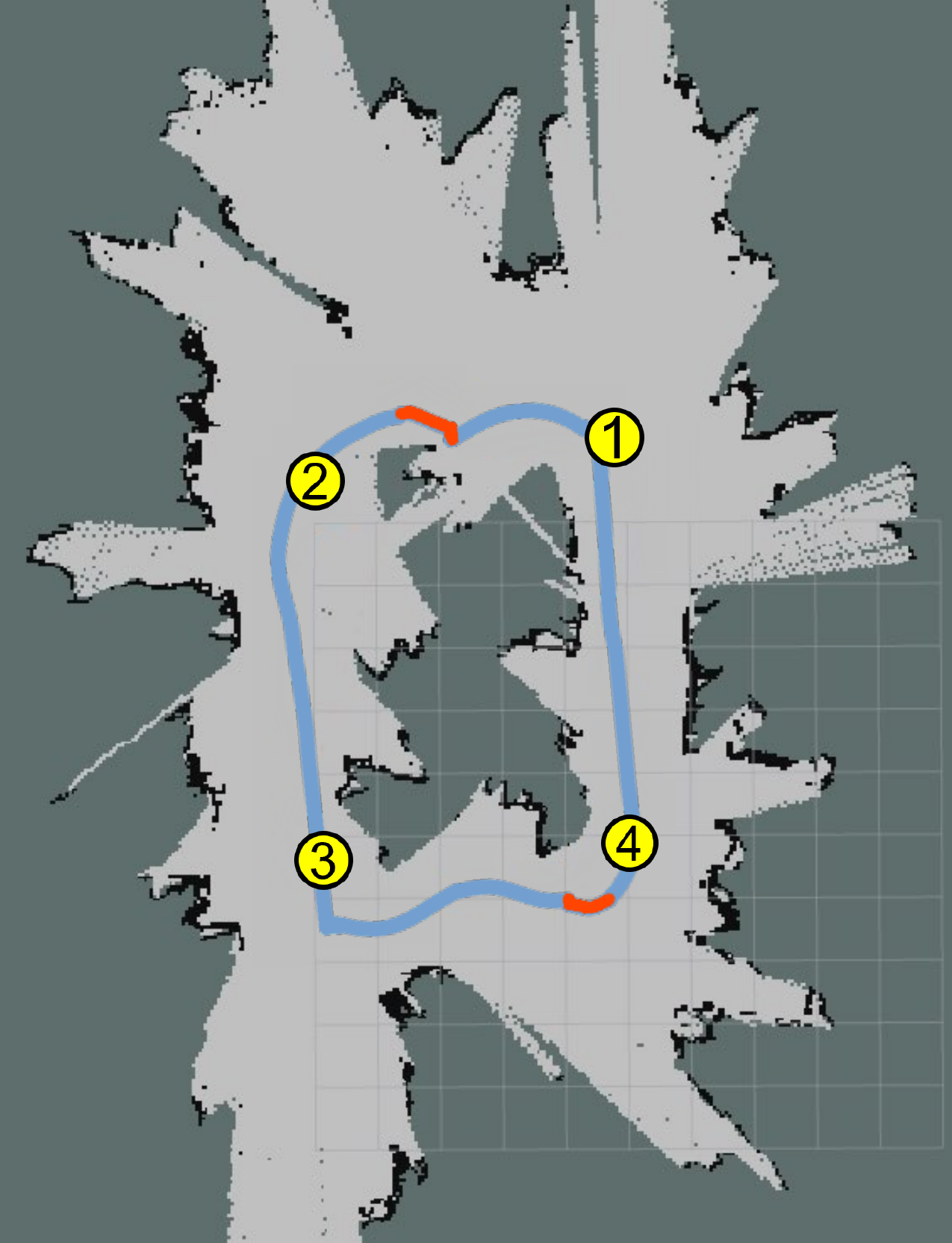}\label{scenario:e}}
\captionsetup{labelfont=bf}
\caption{Qualitative measurement of proposed algorithm and baselines. The solid blue line depicts the ground truth of the trajectory, while the solid red line represents the human-engaged path.}
\label{sim2real:trajectory}
\end{figure*}

\begin{table*}[h]
\renewcommand{\arraystretch}{1.3}
\caption{Quantitative performance of proposed algorithm compared with baselines. The data highlighted in bold denotes better results, i.e., relatively lower mean and variance in terms of the traveling distance and time.}
\label{sim2real}
\begin{adjustbox}{width=1.0\textwidth,center}
\centering

\begin{tabular}{c c c c c c c c}
\hline
\multicolumn{1}{c}{Approach}
&\multicolumn{1}{c}{Goal Position}
&\multicolumn{1}{c}{Avg. Dist.}
&\multicolumn{1}{c}{Var. Dist.}
&\multicolumn{1}{c}{Avg. Time.}
&\multicolumn{1}{c}{Var. Time}
&\multicolumn{1}{c}{Success Rate}
&\multicolumn{1}{c}{Engage Number}\\
\hline
\multirow{4}{*}{GoT-SAC} &1st &6.044 &$\pm$ 0.049 &14.048 &$\pm$ 0.282 &100\% &0\\
                         &2nd &$\mathbf{5.747}$ &$\pm$ $\mathbf{0.110}$ & $\mathbf{11.987}$ &$\pm$ $\mathbf{0.368}$ &100\% &0\\
                         &3rd &$\mathbf{6.171}$ &$\pm$ $\mathbf{0.036}$ &$\mathbf{12.821}$ &$\pm$ $\mathbf{0.076}$ &100\% &0\\
                         &4th &6.360 &$\pm$ 0.113 &13.902 &$\pm$ 1.287 &100\% &0\\
\hline
\multirow{4}{*}{ViT-SAC} &1st &5.958 &$\pm$ 0.036 &15.274 &$\pm$ 0.394 &100\% &0\\
                         &2nd &6.506 &$\pm$ 0.189 &15.432 &$\pm$ 0.741 &100\% &0\\
                         &3rd &6.226 &$\pm$ 0.015 &22.502 &$\pm$ 1.601 &100\% &0\\
                         &4th &7.274 &$\pm$ 0.210 &16.449 &$\pm$ 0.722 &100\% &0\\
\hline
\multirow{4}{*}{ConvNet-SAC\cite{haarnoja2018soft}} &1st &6.150 &$\pm$ 0.294 &17.918 &$\pm$ 2.563 &100\% &4\\
                             &2nd &6.780 &$\pm$ 0.252 &20.735 &$\pm$ 1.519 &100\% &5\\
                             &3rd &6.322 &$\pm$ 0.245 &18.757 &$\pm$ 2.683 &100\% &6\\
                             &4th &5.971 &$\pm$ 0.123 &13.550 &$\pm$ 0.414 &100\% &0\\
\hline
\multirow{4}{*}{MultiModal CIL\cite{xiao2020multimodal}} &1st &5.852 &$\pm$ 0.021 &13.458 &$\pm$ 0.180 &100\% &0\\
                                &2nd &9.868 &$\pm$ 0.480 &57.263 &$\pm$ 6.026 &60\% &5\\
                                &3rd &6.666 &$\pm$ 0.134 &75.121 &$\pm$ 12.019 &20\% &5\\
                                &4th &6.913 &$\pm$ 0.217 &28.893 &$\pm$ 4.475 &100\% &0\\
\hline
\multirow{4}{*}{MoveBase} &1st &5.822 &$\pm$ 0.063 &12.180 &$\pm$ 0.088 &100\% &0\\
                          &2nd &7.070 &$\pm$ 0.529 &23.446 &$\pm$ 9.388 &100\% &4\\
                          &3rd &6.092 &$\pm$ 0.141 &14.010 &$\pm$ 3.122 &100\% &1\\
                          &4th &6.510 &$\pm$ 0.517 &24.098 &$\pm$ 5.442 &100\% &5\\
\hline
\end{tabular}
\end{adjustbox}
\label{sim2real:result}
\end{table*}

\begin{figure}[t]
  \vspace{0.1cm}
  \begin{center}
  \includegraphics[width=3.4in]{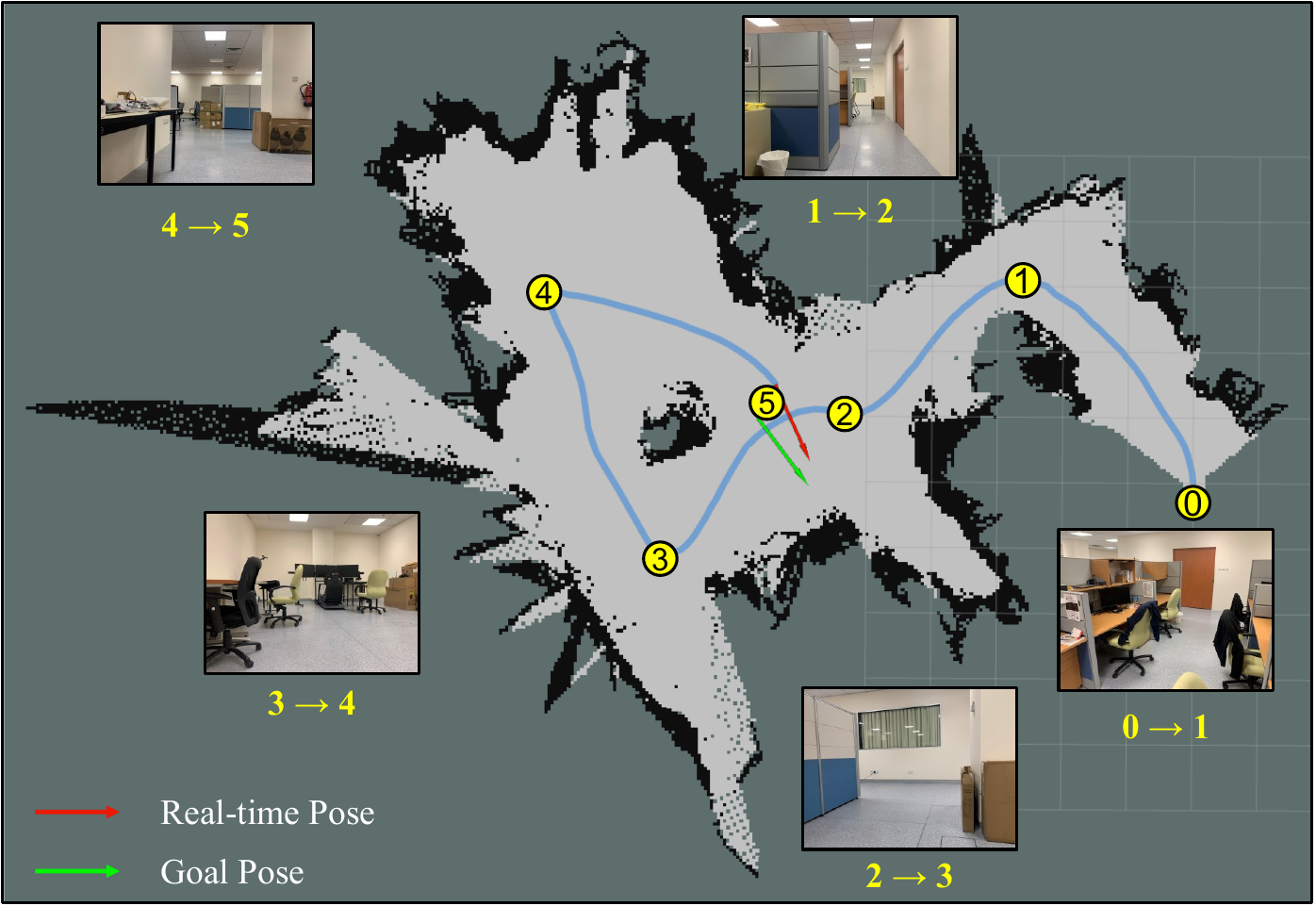}\\
  \caption{Office Environment. The initial and goal positions are labeled by yellow circles, and the performed trajectory is highlighted with the solid blue line.}\label{sim2real:office}
  \end{center}
\end{figure}

It should be mentioned that, unlike the evaluation in the simulation environment, our sim-to-real experiment involves the utilization of visual simultaneous localization and mapping (VSLAM) techniques to provide the ego-pose of the UGV and visualization of the environment. For this purpose, we have implemented the VINS-Fusion framework \cite{qin2018vins}, which has been proven to demonstrate commendable performance within indoor settings. Notably, upon rigorous evaluation using benchmark datasets such as EuRoc \cite{burri2016euroc}, the VINS-Fusion framework has yielded a root square mean error (RSME) of approximately 0.1 meters for the absolute trajectory error (ATE). This level of precision signifies a considerable achievement and holds practical viability for real-world inference applications.

Figure \ref{sim2real:trajectory} illustrates the qualitative measurement of performance for each algorithm. We plot both trajectories from the UGV and the human with two different colors, blue for ground truth and red for human engagement, respectively. As shown in the figure, the GoT-SAC policy performs smooth and collision-free navigation, while the other three algorithms (ConvNet-SAC, MultiModal CIL, and MoveBase) all need human engagement to arrive at the destinations. Surprisingly, we find that ViT-SAC policy also demonstrates an equally excellent performance despite the low average success rate during the evaluation in the simulation environment. It is reasonable since we select the best model for each algorithm for sim-to-real transfer assessment. It may also indicate the significance of the self-attention mechanism for goal-driven autonomous navigation.  

To compare the performance and robustness of the policies in a deeper sight, we employ six statistical metrics for each goal-driven task: the average and variance of traveling distance, average and variance of navigation time, success rate, and engagement number. Especially the successful arrival is determined if the UGV reaches each goal position within one minute, and we actively engage the UGV control once the collision is likely to happen. A detailed quantitative measurement is reported in the Table. \ref{sim2real:result}. It is clear that the GoT-SAC model demonstrates dominant performance and robustness from all the domains compared with other baselines, including ViT-SAC and MoveBase planner. The performance of ViT-SAC is also comparably excellent, besides the longest navigated distance for the fourth destination and high average time for the third goal position. The worst performance is provided by the MultiModal CIL model, whose success rate is only 20\% for reaching the third goal position. We can observe a similar performance from the statistical result of the ConvNet-SAC model in terms of the number of engagements, which is 15 in total. As for the MoveBase planner, the performance highly depends on the local cost-map quality, especially in the turning cases (the second and fourth goal position). For instance, the local cost-map occurs false detection frequently due to limited field of view and occlusion from the obstacles, leading to improper path planning. Overall, both quantitative and qualitative results in this sim-to-real experiment highlight the superiority of the proposed algorithm compared with other baselines, including against SOTA leaning-based approaches and the classic UGV navigation method.

Additionally, our approach is tested in an unknown environment to validate the generalization capability thoroughly. Due to the unstable connection and limitation of hardware, we select an unseen office environment rather than an outdoor space. It is worthwhile to test the generalization and transferability of the proposed algorithm in such an environment since it has a number of corner cases to be addressed, such as planning a collision-free path in narrow corridors, handling unseen obstacles (in terms of shape and color), and performing U-turn operation in order to reach the goal position. Figure \ref{sim2real:office} demonstrates the details of the experiment, where the yellow circle labels the initial and goal positions, and the performed trajectory is highlighted with the solid blue line. In particular, the UGV has to pass a narrow corridor to reach the first two destinations and perform 90-degree-turn and U-turn operations to arrive last two goal positions. Nevertheless, the GoT-SAC model can still approach all five goals without collision or engagement, indicating the excellent generalization capability and transferability of our approach.

\section{Conclusion and Discussions}
This paper presents a Transformer-enabled DRL approach, namely GTRL, to realize efficient goal-driven autonomous navigation. Specifically, we first propose a novel Transformer-based architecture called Goal-guided Transformer (GoT) for the perception to consider the goal information as an input of the scene representation rather than a condition. For the purpose of boosting data efficiency, deep imitation learning is employed to pre-train the GoT. Then, a GoT-enabled soft actor-critic algorithm (GoT-SAC) is instantiated to train the decision policy based on the goal-oriented scene representation. As a result, our approach motivates the scene representation to concentrate mainly on goal-relevant features, which substantially enhances the data efficiency of the DRL learning process, leading to superior navigation performance. Both simulative and sim-to-real transfer experiments confirm our approach's superiority in data efficiency, performance, robustness, and sim-to-real generalization. 

Despite the superior performance demonstrated by the proposed approach compared to other SOTA baselines, instances of failure were observed during the experiments. This can be attributed to the ground reflection resulting from the lighting conditions. Specifically, under strong lighting, the ground reflection introduces a substantial deviation between the original appearance and the RGB images captured by the fisheye camera, consequently leading to unfavorable navigation performance. Furthermore, the current approach exhibits degraded navigation performance when encountering dynamic obstacles, particularly pedestrians wearing clothes in a similar color to the ground, since it has never encountered such obstacles during the training phase.

In light of this, our future objective is to transfer the navigation environment from indoor to outdoor and incorporate a more diverse range of input modalities for our model. The navigation task becomes significantly more complex in the outdoor environment as the UGV must handle varying lighting conditions and highly dynamic pedestrians. To address these challenges, the adoption of multiple input modalities, such as occupancy flow or segmentation images, could be considered to perform interactive fusion for segmenting the drivable area explicitly, filtering out irrelevant information and thereby enhancing subsequent outdoor navigation performance. Moreover, we intend to incorporate abnormal detection and re-localization functions to mitigate the impact of position errors on navigation performance in the future. These mechanisms will help identify and handle situations where position errors exceed acceptable thresholds, maintaining a reliable and consistent navigation performance even in challenging scenarios.

\bibliographystyle{IEEEtran}
\bibliography{IEEEabrv,mybib}

\vfill

\end{document}